\documentclass[runningheads]{llncs}
\usepackage{graphicx}
\usepackage{amsmath,amssymb} % define this before the line numbering.
\usepackage{color}
\usepackage[width=122mm,left=12mm,paperwidth=146mm,height=193mm,top=12mm,paperheight=217mm]{geometry}
\usepackage{float}
\usepackage[caption=false,font=small]{subfig}

\begin{document}
% \linenumbers
\pagestyle{headings}
\mainmatter

\title{ArticulatedFusion: Real-time Reconstruction of Motion, Geometry and Segmentation Using a Single Depth Camera} 
% Replace with your title

\titlerunning{ArticulatedFusion}
% Replace with a meaningful short version of your title

\authorrunning{C. Li, Z. Zhao and X. Guo}
% Replace with shorter version of the author list. If there are more authors than fits a line, please use A. Author et al.

\author{Chao Li \and Zheheng Zhao \and Xiaohu Guo}

%Please write out author names in full in the paper, i.e. full given and family names. 
%If any authors have names that can be parsed into FirstName LastName in multiple ways, please include the correct parsing, in a comment to the volume editors:
%\index{Lastnames, Firstnames}
%(Do not uncomment it, because you may introduce extra index items if you do that, we will use scripts for introducing index entries...)

\institute{Department of Computer Science,\\
	The University of Texas at Dallas\\
	\email{ \{Chao.Li2, Zheheng.Zhao, xguo\}@utdallas.edu}
}

\maketitle

\begin{abstract}
	This paper proposes a real-time dynamic scene reconstruction method capable of reproducing the motion, geometry, and segmentation simultaneously given live depth stream from a single RGB-D camera. Our approach fuses geometry frame by frame and uses a segmentation-enhanced node graph structure to drive the deformation of geometry in registration step. A two-level node motion optimization is proposed. The optimization space of node motions and the range of physically-plausible deformations are largely reduced by taking advantage of the articulated motion prior, which is solved by an efficient node graph segmentation method. Compared to previous fusion-based dynamic scene reconstruction methods, our experiments show robust and improved reconstruction results for tangential and occluded motions. 

\end{abstract}

\section{Introduction}
Dynamic scene reconstruction is a very important topic for digital world building.
It includes capturing and reproducing geometry, appearance, motion, and skeleton, which enables more realistic rendering for VR/AR scenarios like Holoportation \cite{dou2016fusion4d}. An example is that the reconstructed geometry can be directly used for a virtual scene, and the articulated motion can be retargeted to new models to generate new animations, making scene production more efficient.

Although many efforts have been devoted to this research field, the problem remains challenging due to extraordinarily large solution space but real-time rendering requirements for VR/AR applications. Recently, volumetric depth fusion methods for dynamic scene reconstruction, such as DynamicFusion \cite{newcombe2015}, VolumeDeform \cite{innmann2016volumedeform}, Fusion4D \cite{dou2016fusion4d} and albedo based fusion \cite{guo2017} open a new gate for people in this field. This type of method enables quality improvements over temporal reconstruction models in terms of both accuracy and completeness of the surface geometry. Among all these works, fusion methods by a single depth camera \cite{newcombe2015,innmann2016volumedeform} are more promising for popularization, because of their low cost and easy setup. However, this group of methods still faces some challenging issues, like high occlusion from the single view, limited computational resource to achieve real-time performance, and no geometry/skeleton prior knowledge, and thus are restricted to limited motions. DoubleFusion \cite{DoubleFusion} can reconstruct both the inner body and outer surface for faster motions by adding body template as prior knowledge. Later, KillingFusion \cite{slavcheva2017cvpr} and SobolevFusion \cite{slavcheva2018cvpr} is proposed to reconstruct dynamic scenes with topology changes and fast inter-frame motions.

DynamicFusion is the pioneering work acheiving template-less non-rigid reconstruction in real time from single depth camera. However, its robustness can be significantly improved by utilizing skeleton prior, as been shown in work of BodyFusion \cite{yu12bodyfusion}. In this paper, we propose to add articulated motion prior into the depth fusion system. Our method contributes to this field by pushing the limitation from skeleton-prior-based methods to skeleton-less ones. The motions of many objects in our world including human motion follows articulated structures. Thus, articulated motions can be represented by skeleton/cluster-based motion and can be extracted from non-rigid motion as a prior. Our self-adaption segmentation inherits the rigid feature of traditional skeleton structure while does not require any pre-defined skeleton. The segmentation constrains all nodes labeled to the same segment having transformation as close as possible and can reduce the solution space of the optimization problem. Therefore, the self-adapted segmentation can result in better reconstruction results.

%Using such a prior can constrain any large non-rigid deformation to be consistent with the skeletal joints or overlapping regions, thereby reducing the optimization space and the range of physically-plausible deformations. Based on these observations, we propose segmentation-involved approach to improve the reconstruction quality of dynamic scenes for both human and non-human objects.

Our method iteratively optimizes the motion field of a node graph and its segmentation, which helps each other to get a better reconstruction performance. Integrating the articulated motion prior into the reconstruction framework assists in the non-rigid surface registration and geometry fusion, while surface registration results improve the quality of segmentation and its reconstructed motion. Although the advantages of such unification is obvious, in practice, designing a real-time algorithm to take advantage of both merits of these two aspects is still an unstudied problem, especially on how to segment a node graph based on its motion trajectory in real-time. We have carefully designed our ArticulatedFusion system, to achieve simultaneous reconstruction of motion, geometry, and segmentation in real-time, given a single depth video input. The contributions in this paper are as follows:
\begin{enumerate}
	\item We present ArticulatedFusion, a system that involves registration, segmentation, and fusion, and enables real-time reconstruction of motion, geometry, and segmentation for dynamic scenes of human and non-human subjects. 
	\item A two-level registration method which can narrow down the optimization solution space, and result in better reconstructed motions in many challenging cases, with the help of node graph segmentation.
	\item A novel real-time segmentation method to solve the clustering of a set of deformed nodes based on their motion trajectories by merging and swapping operations.
\end{enumerate}

%-------------------------------------------------------------------------

\section{Related Work}

The most popular dynamic 3D scene reconstruction method is to use a predefined model or skeleton as prior knowledge. Most of these methods focus on the reconstruction of human body parts such as face \cite{li2013realtime,cao20133d}, hands \cite{tagliasacchi2015robust,tkach_siga16}, and body \cite{pons2010multisensor,vlasic2008articulated}. Other techniques are proposed to reconstruct general objects by using a pre-scanned geometry~\cite{li2009,zollhofer2014real} as a template instead of predefined models. 

To further eliminate the dependency on geometry priors, some template-less methods were proposed to utilize more advanced structure to merge and store geometry information across the motion sequence. Wand et al. \cite{wand2009efficient} proposed an algorithm to align and merge pairs of adjacent frames in a hierarchical fashion to gradually build the template shape. Recently, fine 3D models have been reconstructed without any shape priors by gradually fusing multi-frame depth images from a single view depth camera \cite{newcombe2011kinectfusion,newcombe2015,innmann2016volumedeform,dou2016fusion4d}. Innmann et al. \cite{innmann2016volumedeform} proposed to add SIFT features to the ICP registration framework, thereby improving the accuracy of motion reconstruction. 

Our method is partly inspired by Pekelny and Gotsman's method \cite{pekelny2008}, However their method
requires the user to manually segment a range scan in advance, whereas we automatically solve for the segmentation in real-time. Chang and Zwicker's method \cite{chang2011} is also lack of real data of human motions and takes much time to reconstruct for each frame. Tzionas and Gall's recent work \cite{Tzionas:ECCVw:2016} introduces an algorithm to build rigged models of articulated objects from depth data of a single camera. But it requires to pre-scan the target object as the geometry prior knowledge.

Guo et al. ~\cite{guo2015} proposes an $L_0$ regularizer to constrain local non-rigid deformation only on joints of articulated motion, which reduces the solution space and yields a physically plausible and robust deformation. However, our method is designed to achieve real-time performance while their method requires around 60s for the $L_0$ optimization of each frame~\cite{guoTVCG2017}.
Ours directly solves the segmentation of human
body in the proposed energy function while theirs implicitly involves the articulated motion property in an $L_0$ regularizer.
Their method also needs a pre-scaned shape as a template.
Yu et al.’s method \cite{yu12bodyfusion} is the one most related to our work,
but it requires the skeleton information of the first frame as
initialization while our method does not need any prior information.
Our method can estimate the segmentation of dynamic scene during the reconstruction process. Therefore, it also works for non-human objects where a predefined skeleton is not available, as illustrated in Fig.~\ref{fig:nonHuman} and Fig.~\ref{fig:nonHumanCom}.
There is also a rich body of work proposed on articulated decomposition of animated mesh sequences~\cite{james2005skinning,le2012smooth}. Both of these methods can only work on animated sequences with fixed mesh connectivity, and cannot meet our real-time reconstruction requirement. 

%-------------------------------------------------------------------------

\section{Overview}
\label{sec:overview}

\begin{figure*}[t]
	\centering
	\includegraphics[width=1.0\textwidth]{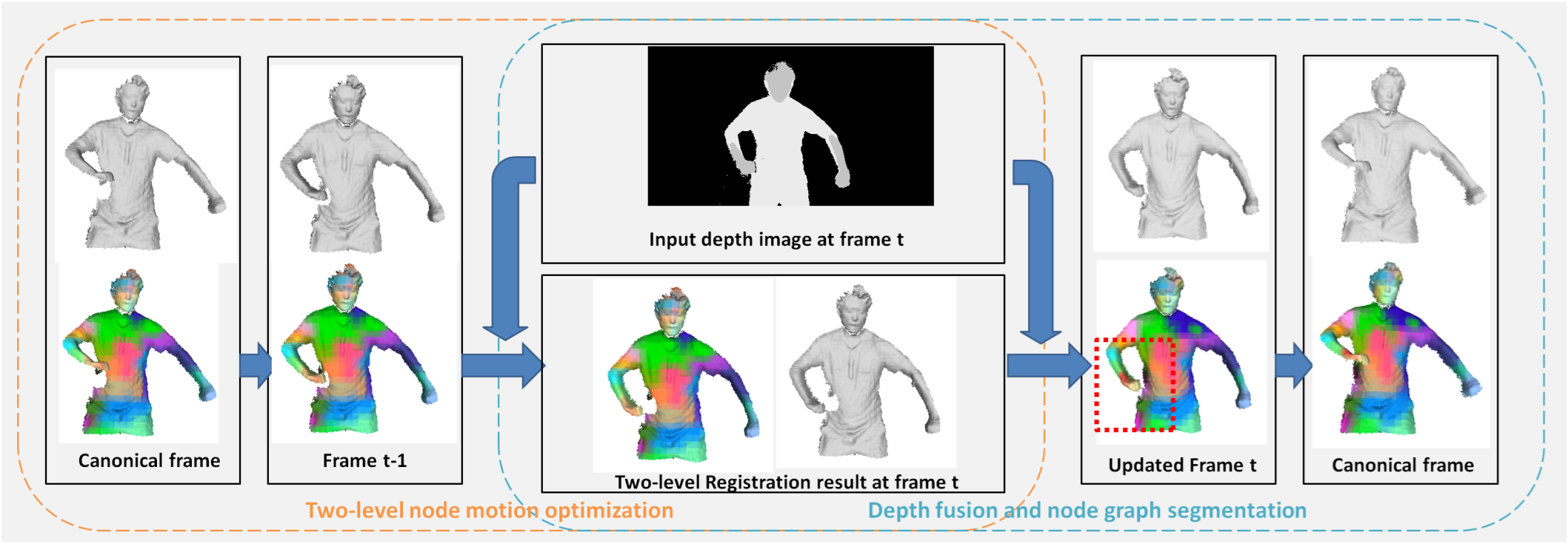}\\
	\caption{Overview of our pipeline. The orange box represents our two-level node motion 
    optimization, and the blue box represents fusion of depth and node graph segmentation. }
	\label{fig:flow}
\end{figure*}

Fig. \ref{fig:flow} illustrates the pipeline of processing one frame given the geometry, motion and segmentation reconstructed from earlier frames. 
\cite{newcombe2015,innmann2016volumedeform,guo2017}, our system runs in a frame-by-frame manner.
Two main data structures are used in our system. The geometry is represented in a volume with the Truncated Signed Distance Function (TSDF), while the segmentation and motions are defined in an embedded graph of controlling nodes similar to DynamicFusion \cite{newcombe2015}. 

The first frame is selected as the \textit{canonical frame}. The first step of our system is \textit{two-level node motion optimization} (Sec. \ref{subsec:reg}). In this step, motions of controlling nodes from the canonical frame to the current frame are estimated. This is achieved by first warping a mesh using reconstructed motion and segmentation from earlier frames, and followed by solving a two-level optimization problem to fit this mesh with the current depth image. The mesh is extracted from the TSDF volume by marching cube algorithm \cite{lorensen1987marching}. The first level of our node motion optimization is run on each segmented cluster, thus can reduce the solution space and make optimization converging faster. The second level of optimization is run on each individual node, so it can keep track of the high-frequency details of the target object. The depth is fused into the TSDF volume to obtain a new integrated geometry (Sec. \ref{subsec:fuse}). The final step is \textit{node graph segmentation}, in which nodes are segmented by our novel clustering method to minimize the error between the articulated cluster deformation of nodes and their non-rigid deformation (Sec. \ref{subsec:seg}). 
This segmentation makes the node motion estimation of next frame to perform better than employing non-rigid estimation only.

%-------------------------------------------------------------------------

\section{Method}
\subsection{Preliminaries and Initialization}
\label{subsec:init}
Only a single depth camera is used to capture the depth information in our system. 
The input to our pipeline is a depth image sequence $\{\mathcal{D}^{t}\}$. 
The output of our pipeline includes a fused geometry $\mathcal{V}$ of the target object, the embedded graph segmentation $\mathcal{C}$, and the two-level warping field $\{\mathcal{W}^{t}\}$, where $\mathcal{W}^{t}$ represents the non-rigid node motion from the canonical frame to each live frame $t$. The TSDF volume and level-two warping field in our system is the same as those described in DynamicFusion \cite{newcombe2015}.

For the first frame, we directly integrate the depth information into the canonical TSDF volume, extract a triangular mesh $\mathcal{M}$ from the canonical volume using the marching cube algorithm, uniformly sample deformation nodes on the mesh and construct a node graph to describe the non-rigid deformation. To search for nearest-neighboring nodes, we also create a dense k-NN field in the canonical volume. Because our segmentation method is based on the motion trajectory from canonical frame to a live frame, we cannot get a segmentation result for the first frame. Therefore, we employ the non-rigid registration method of DynamicFusion \cite{newcombe2015} to align the mesh to the second frame.

\subsection{Registration}
\label{subsec:reg}
As mentioned above, the first step of our system is to fit the canonical mesh $\mathcal{M}$ to the depth image $\mathcal{D}^{t}$ of live frame $t$. We have the current mesh $\mathcal{M}$ (obtained by fusing the depth from earlier frames), the segmentation $\mathcal{C}$, and the motion field $\mathcal{W}^{t-1}$. Using the newly captured depth in frame $t$, the algorithm presented in this section estimates $\mathcal{W}^{t}$ to fit $\mathcal{M}$ with $\mathcal{D}^{t}$. For this purpose, we propose a two-level optimization framework based on Linear Blend Skinning (LBS) model and node graph motion representation. The optimization is solved by minimizing the following energy function first in LBS model and then in node graph model:
\small
\begin{equation}
\label{eqn:total_energy}
E_{total}(\mathcal{W}^{t}) = \omega_{f} E_{fit} + \omega_{r} E_{reg},
\end{equation}
\normalsize
where $E_{fit}$ is the data term to minimize the fitting error between deformed vertex and its corresponding point on depth image, and $E_{reg}$ regularizes the motion to be locally as rigid as possible. $\omega_{f}$ and $\omega_{r}$ are controlling weights to balance the influence of two energy terms. In all of our experiments, we set $\omega_{f}=1.0$ and $\omega_{r}=10.0$

Before solving the energy function, we build the two-level deformation model based on the node graph and its segmentation by defining the following skinning weight for each vertex $\mathbf{v}_{i}$ on mesh $\mathcal{M}$:
\small
\begin{equation}
\label{eqn:weight}
\mathbf{w}_{i}^{(l)} = \left\{
\begin{array}{lll}
\frac{1}{\Lambda}\sum_{j=1}^{k}\lambda_{i,j}\mathbf{g}_{j} & &l=1,\\
\frac{1}{\Lambda}\sum_{j=1}^{k}\lambda_{i,j}\mathbf{h}_{j} & &l=2,
\end{array}
\right.
\end{equation}
\normalsize
where $l$ denotes the level, and $\lambda_{i,j}$ is the weight describing the influence of the $j$-th node $\mathbf{x}_{j}$ on vertex $\mathbf{v}_{i}$ and is defined as $\lambda_{i,j} = exp\left( -\|\mathbf{v}_{i}-\mathbf{x}_{j}\|^{2}_{2}/ \left( 2 \sigma_{j} \right) ^{2}\right) $. $\Lambda$ is a normalization coefficient, the summation of all spatial weights $\lambda_{i,j}$ for the same $i$. Here, $\sigma_{j}$ is the given influence radius of controlling node $\mathbf{x}_{j}$. When level $l=1$, $\mathbf{g}_{j}=\left( g_{j,1}, g_{j,2},...,g_{j,m}\right)$ is the binding of controlling node $\mathbf{x}_{j}$ to $m$ clusters. Because each node only belongs to one cluster, only one element of $\mathbf{g}_{j}$ is $1$ and all other elements are $0$. $\mathbf{w}_{i}^{(1)}=\left( w_{i,1}^{(1)}, w_{i,2}^{(1)},...,w_{i,m}^{(1)} \right)$ includes the skinning weights of vertex $\mathbf{v}_{i}$ w.r.t. $m$ clusters. When level $l=2$, $\mathbf{h}_{j}=\left( h_{j,1}, h_{j,2},...,h_{j,k}\right) $ is the binding of $\mathbf{v}_{i}$'s neighboring node $\mathbf{x}_{j}$ to itself. Thus only $h_{j,j}$ is $1$ and all other elements are $0$. $\mathbf{w}_{i}^{(2)}=\left( w_{i,1}^{(2)}, w_{i,2}^{(2)},...,w_{i,k}^{(2)} \right)$ includes the skinning weight of vertex $\mathbf{v}_{i}$ w.r.t. its k-NN controlling nodes.

The fitting term $E_{fit}$ represents the point-to-plane energy, as follows:
\small
\begin{equation}
\label{eqn:fit_term}
E_{fit}(\mathcal{W}^{t}) = \sum_{i} \left(\mathbf{n}^{\top}_{\mathbf{u}_{i}^{t}}\left( \mathbf{\hat{v}}_{i}-\mathbf{u}_{i}^{t}\right)  \right) ^{2},
\end{equation}
\normalsize
where $\mathbf{\hat{v}}_{i}$ is the transformed vertex defined by the formula: 
\small
\begin{equation}
\mathbf{\hat{v}}_{i} = \sum_{j}w_{i,j}^{(l)}\left(\mathbf{R}_{j}^{t}\mathbf{v}_{i}+\mathbf{t}_{j}^{t}\right).
\end{equation}
\normalsize
Here $\mathbf{v}_{i}$ is a vertex on $\mathcal{M}$, and $\{\mathbf{R}_{j}^{t},\mathbf{t}_{j}^{t}\}$ are the unknown rotation and translation of either the $j$-th cluster (level $l=1$) or the $j$-th node (level $l=2$), which will be solved during the optimization process. $\mathbf{u}_{i}^{t}$ is the corresponding 3D point on depth frame $D^{t}$ for $\mathbf{v}_{i}$, and $\mathbf{n}_{\mathbf{u}_{i}^{t}}$ represents its normal. To obtain the pair of such correspondences, we render the deformed mesh $\mathcal{M}$ with the current warping field to exclude occluded vertices and project visible vertices onto the screen space of $D^{t}$ . Then we look up the corresponding pixel with the same coordinates. For vertices lying on the silhouette of projected 2D image, we employ Tagliasacchi et al.'s method \cite{tagliasacchi2015robust} -- using 2D Distance Transform (DT) to locate the corresponding pixel and back-projecting it to 3D camera space. This correspondence search mechanism can guarantee better convergence when meeting large deformations in the direction perpendicular to the screen space (tangential motions) between two adjacent frames. 
\begin{figure}
	\centering
	\subfloat[DT of the 2nd frame ]{\includegraphics[width=0.4\textwidth]{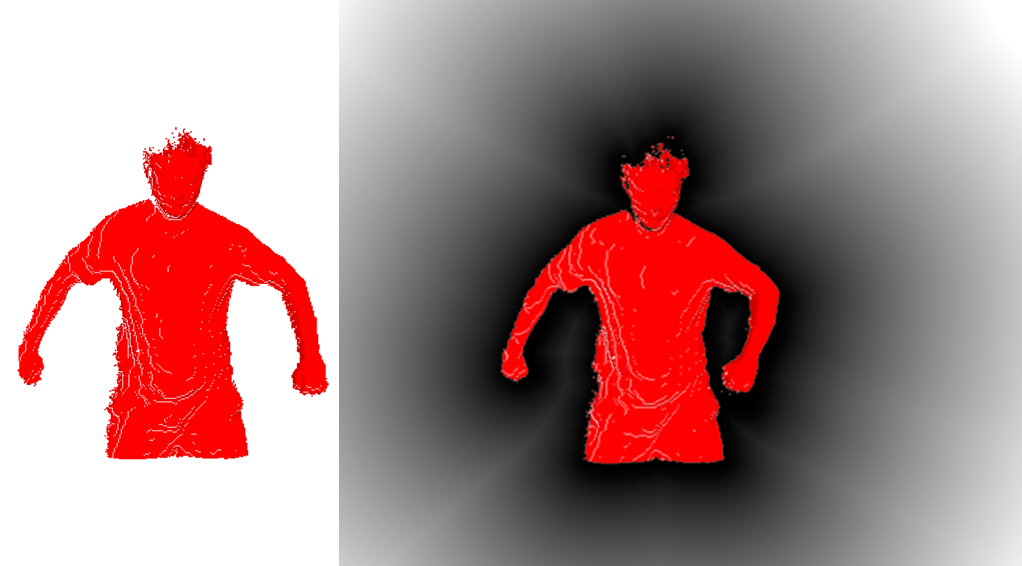}}
	~~\subfloat[Without DT]{\includegraphics[width=0.25\textwidth]{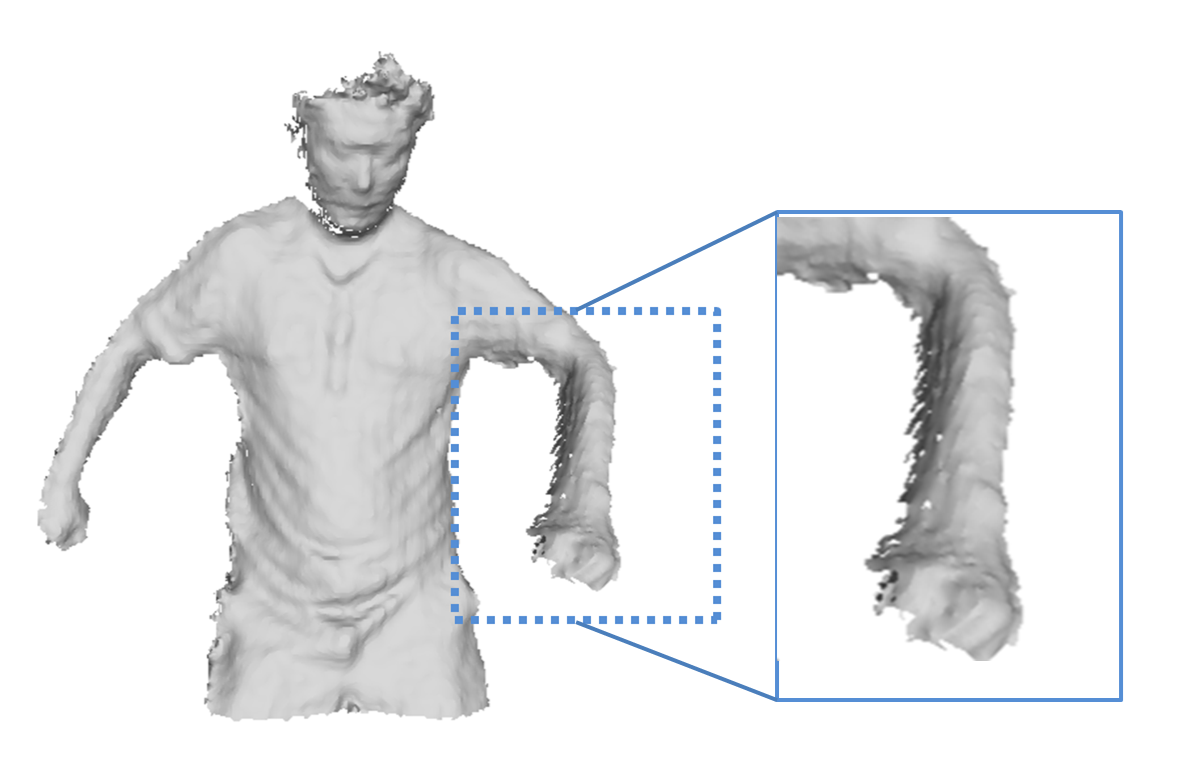}}
	~~\subfloat[With DT]{\includegraphics[width=0.25\textwidth]{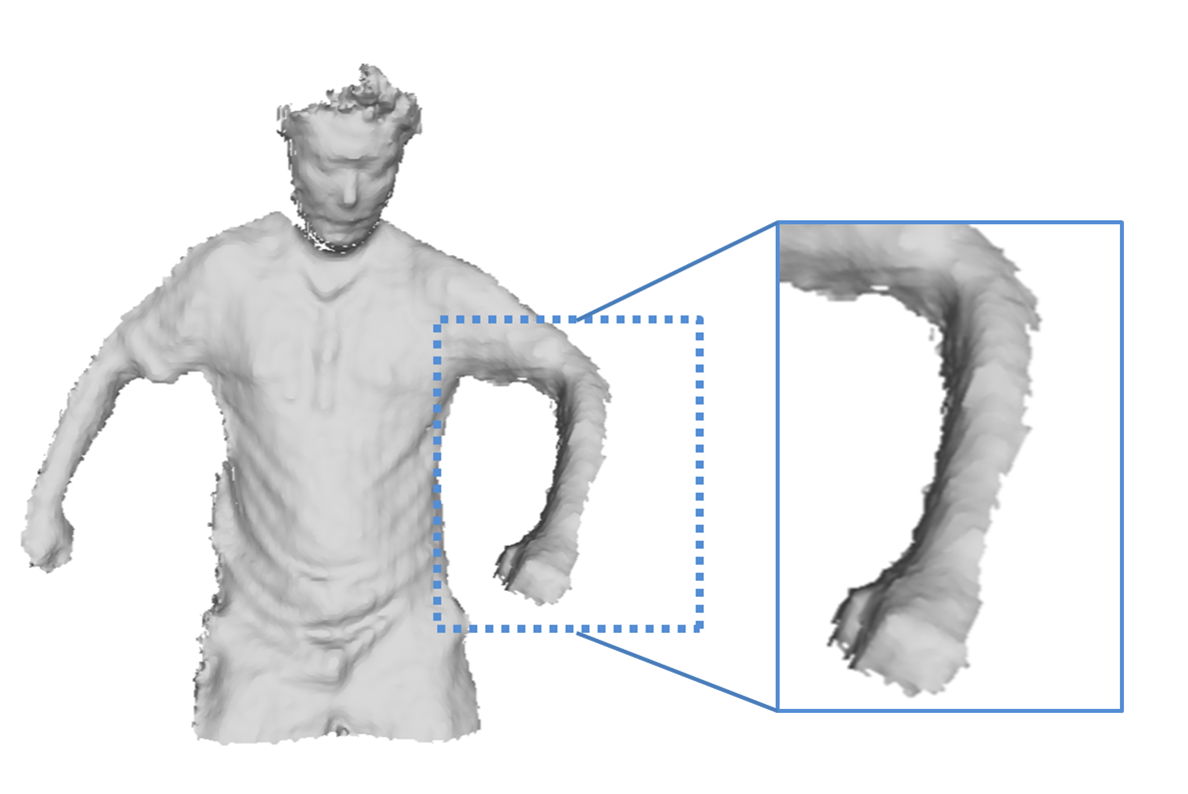}}
	\caption{Tracking results comparison from one frame to its next frame without and with Distance Transform (DT) correspondences.}
	\label{fig:distTrans}
\end{figure}
Fig.\ref{fig:distTrans} shows a comparison of results with and without distance transform correspondences. 
Fig.\ref{fig:distTrans} (a) are point clouds from two adjacent frames. The subfigure on the right illustrates the computed distance transform based on depth image contour.
Fig.\ref{fig:distTrans} (b) represents the tracking reconstruction result without using distance transform correspondences for silhouette points while Fig. \ref{fig:distTrans} (c) represents the result with distance transform correspondences search which is converged better than the one in Fig. \ref{fig:distTrans} (b).

The regularity term $E_{reg}$ is an as-rigid-as-possible constraint:
\small
\begin{multline}
\label{eqn:reg_term}
E_{reg}(\mathcal{W}^{t}) = \sum_{j_1}\sum_{j_2\in \mathcal{N}(j_1)} \alpha^{(l)}(\mathbf{g}_{j_1},\mathbf{g}_{j_2}) \cdot 
\|\mathbf{R}_{j_1}^{t}\mathbf{x}_{j_2}+\mathbf{t}_{j_1}^{t}-\mathbf{R}_{j_2}^{t}\mathbf{x}_{j_2}-\mathbf{t}_{j_2}^{t}\|^{2},
\end{multline}
\normalsize
where $\mathcal{N}(j_1)$ denotes the set of neighboring nodes of the $j_1$-th node. $\alpha^{(l)}(\mathbf{g}_{j_1},\mathbf{g}_{j_2})$ is a clustering-awareness weight. In level $l=1$, $\alpha^{(1)}(\mathbf{g}_{j_1},\mathbf{g}_{j_2})=1$ when the $j_1$-th node and the $j_2$-th node belong to the same cluster, and $\alpha^{(1)}(\mathbf{g}_{j_1},\mathbf{g}_{j_2})=0$ otherwise. In level $l=2$, $\alpha^{(2)}(\mathbf{g}_{j_1},\mathbf{g}_{j_2})$ is always equal to 1. This regularization term is important to ensure that all vertices will move with the visible regions as rigidly as possible if some object regions are occluded due to our single-camera capture environment.

The minimization of Eq. (\ref{eqn:total_energy}) is a nonlinear problem. In level $l=1$, we solve the transformations of each cluster, while in level $l=2$, we solve the transformations of each node. Both levels are solved through Gauss-Newton iterations. In each iteration, the problem is linearized around the transformations from the previous iteration: $\mathbf{J}^{\top}\mathbf{J}\mathbf{\hat{x}} = \mathbf{J}^{\top}\mathbf{f}$, where $\mathbf{J}$ is the Jacobian of function $\mathbf{f}(\hat{x})$ from the energy decomposition: $E_{total}(\hat{x})=\mathbf{f}(\hat{x})^{\top}\mathbf{f}(\hat{x})$. Then, a linear system is solved to obtain the updated transformations of $\mathbf{\hat{x}}$ for the current iteration with the twist representation \cite{murray1994mathematical} to represent the 6D motion parameters of each cluster or node. In order to meet the real-time requirement, we use the same method as in Fusion4D \cite{dou2016fusion4d}: $\mathbf{J^{\top}\mathbf{J}}$ and $\mathbf{J}^{\top}\mathbf{f}$ is constructed on GPU, and then Preconditioned Conjugate Gradient (PCG) method is employed to solve the transformations.

\subsection{Depth Fusion}
\label{subsec:fuse}
After solving for the deformation of each node, we integrate the depth information into the TSDF volume of canonical frame and uniformly sample the newly added surface to update the nodes \cite{newcombe2015}. However, this integration method may result in issues due to voxel collision: if several voxels are warped to the same position in the live frame, then the TSDF of all these voxels will be updated. To resolve this ambiguity, we modify the method presented in Fusion4D \cite{dou2016fusion4d} to a stricter strategy. If two or more voxels in the canonical frame are warped to the same position, we reject their TSDF integration. This method avoids the generation of erroneous surfaces due to voxel collisions.

\subsection{Segmentation}
\label{subsec:seg}
The optimal articulated clustering of node graph  $\mathcal{C}=\{C_{n}\}$ can be solved based on the motion trajectory from the canonical frame to live frame $t$. We assume that each cluster is associated with a rigid transformation $\{\mathbf{R}_{n}^{t}, \mathbf{t}_{n}^{t}\}$. The following energy function measures the error between rigidly transformed node positions to their non-rigidly warped positions in live frame $t$:
\small
\begin{equation}
\label{eqn:decompose}
E_{seg}=\sum_{n=1}^{m}\sum_{\mathbf{x}\in {C_{n}}}{\|\mathbf{R}_{n}^{t}\mathbf{x}+\mathbf{t}_{n}^{t}-\mathbf{y}^{t}\|}^{2},
\end{equation}
\normalsize
where $t$ is the index of the live frame, $n$ is the index of clusters, $m$ is the total number of clusters,
$\mathbf{x}$ is position of a node in the canonical frame and $\mathbf{y}^{t}$ is its corresponding node position after being warped into frame $t$.
$\mathbf{x}$ and $\mathbf{y}^{t}$ have one-to-one correspondence because $\mathbf{y}^{t}$ are all deformed from the canonical frame.

The minimization of Eq. (\ref{eqn:decompose}) implicitly includes the information of the motion trajectory -- nodes with similar motions will be merged into the same cluster. By using our following method, the unknown clustering $\{C_{n}\}$ and per-cluster transformation $\{\mathbf{R}_{n}^{t},\mathbf{t}_{n}^{t}\}$ can be solved simultaneously and efficiently. Although they are correlated, we find that $\{\mathbf{R}_{n}^{t},\mathbf{t}_{n}^{t}\}$ has a closed-form solution for fixed clustering in Eq. (\ref{eqn:decompose}) \cite{Horn87closed-form,SORKINE2007}. In this paper, we employ the merging and swapping idea as proposed by Cai et al. \cite{cai2016CGF,cai2017TVCG} to solve for $\{C_{n}\}$ and $\{\mathbf{R}_{n}^{t},\mathbf{t}_{n}^{t}\}$ simultaneously.

Now we formulate the optimal clustering by minimizing the energy of Eq. (\ref{eqn:decompose}) while keeping their rigid transformation $\{\mathbf{R}_{n}^{t},\mathbf{t}_{n}^{t}\}$ fixed:
\small
\begin{equation}
\label{eq:clustering}
{\{C_{n}\}}^m_{n=1}=\min_{C_{n}}\sum_{n=1}^{m} \sum_{\mathbf{x} \in C_{n}}\|\mathbf{R}_{n}^{t}\mathbf{x}+\mathbf{t}_{n}^{t}-\mathbf{y}^{t} \|^{2}.
\end{equation}
\normalsize
For each cluster $C_{n}$, we define its centroid in the canonical frame as $\mathbf{c}_{n}$:
\small
\begin{equation}
\label{eq:centroid1}
\mathbf{c}_{n}=\frac{\sum_{\mathbf{x} \in C_{n}}\mathbf{x}}{\sum_{\mathbf{x} \in C_{n}} 1},
\end{equation}
\normalsize
and so is its corresponding vertex centroid $\mathbf{c}_{n}^{t}$ in live frame $t$.
Then Eq. (\ref{eq:clustering}) can be rewritten by applying the closed-form solution of $\{\mathbf{R}_{n}^{t},\mathbf{t}_{n}^{t}\}$:
\small
\begin{equation}
\label{eq:optimal-clustering}
{\{C_{n}\}}^m_{n=1}=\min_{C_{n}}\sum_{n=1}^{m}E^{*}(C_{n}),
\end{equation}
\normalsize
where:
\small
\begin{multline}
\label{eq:optimal-extend-energy}
E^{*}(C_{n})=\sum_{\mathbf{x} \in C_{n}}[(\mathbf{x}-\mathbf{c}_{n})^{\top}(\mathbf{x}-\mathbf{c}_{n})
+(\mathbf{y}^{t}-\mathbf{c}^{t}_{n})^{\top}(\mathbf{y}^{t}-\mathbf{c}^{t}_{n})]-2\sum_{q=1}^{3}\sigma_{nq}^{t},
\end{multline}
\normalsize
and $\sigma_{nq}^{t}$ is the singular value of cross covariance matrix $\mathbf{A}^{t}(C_{n})$:
\small
\begin{equation}
\label{eq:covariance}
\mathbf{A}^{t}(C_{n})=\sum_{\mathbf{x} \in C_{n}}(\mathbf{x}-\mathbf{c}_{n})(\mathbf{y}^{t}-\mathbf{c}^{t}_{n})^{\top}.
\end{equation}
\normalsize

Eq. (\ref{eq:optimal-clustering}) can be solved in two stages: initial clustering by merging operations, and clustering optimization by swapping operations.

\textbf{Initial Clustering by Merging Operations:}
Inspired by the surface simplification idea of Cai et al. \cite{cai2017TVCG}, we define merging operations to partition the nodes of the canonical frame into \emph{m} clusters as initialization. It will result in a good initial clustering for the next stage of swapping-based optimization.

\begin{figure}[t]
	\centering
	\subfloat[]{\includegraphics[width=1.0in]{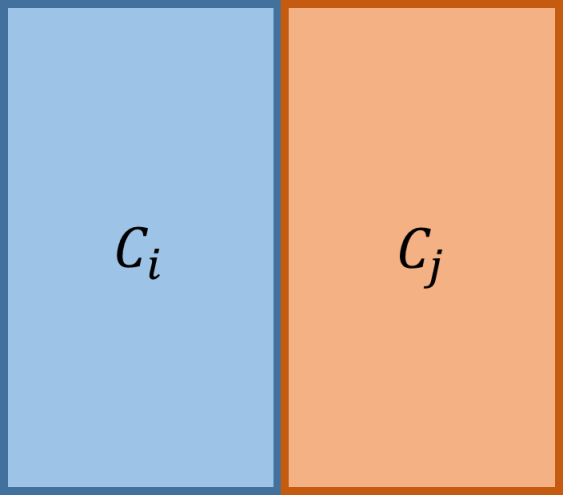}}
	~\subfloat[]{\includegraphics[width=1.0in]{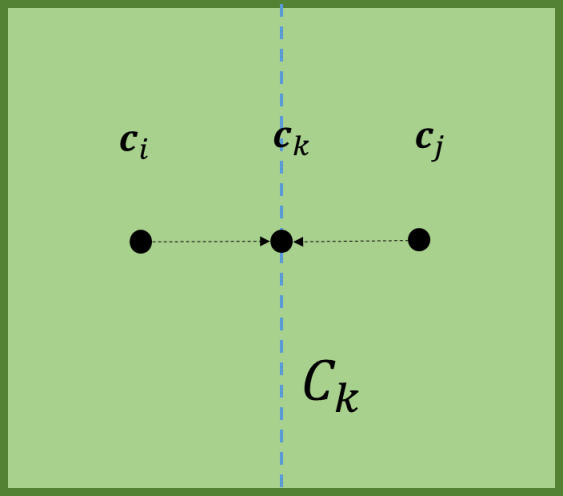}}	
	~\subfloat[]{\includegraphics[width=0.8in]{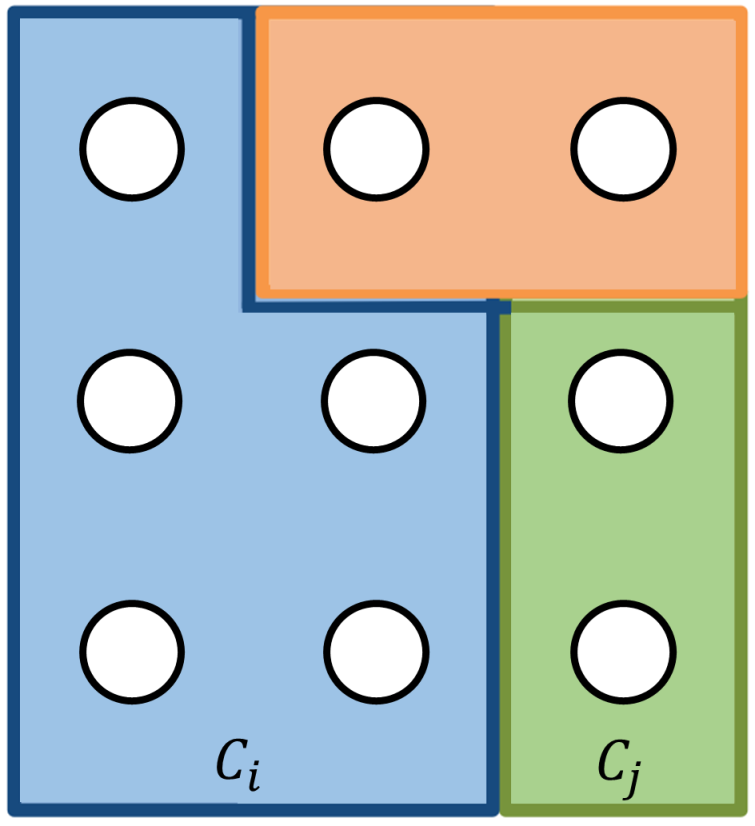}}
	~\subfloat[]{\includegraphics[width=0.8in]{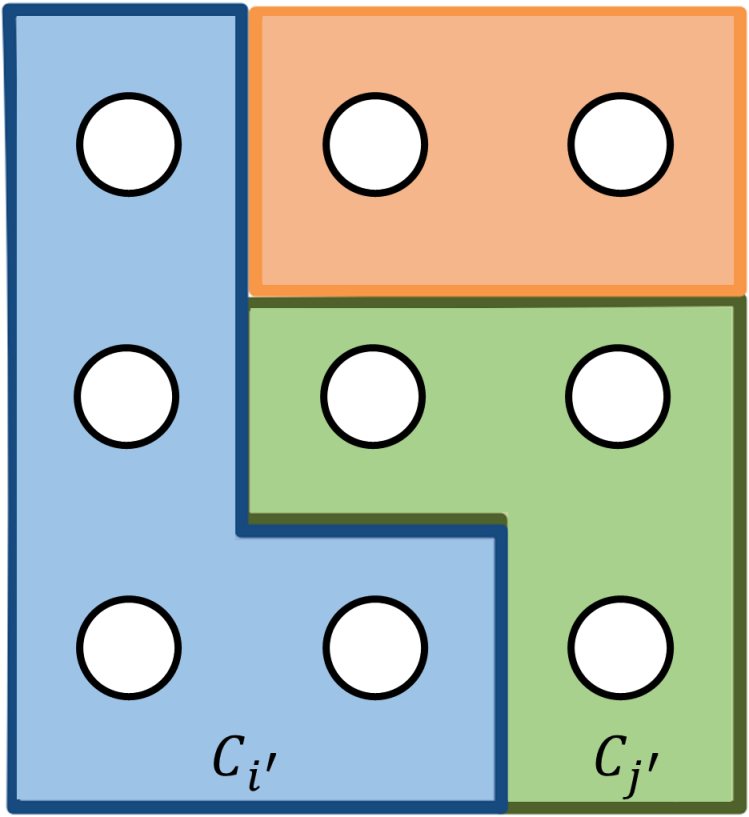}}
	
	\caption{Merging and swapping operation for a pair of clusters.
		$C_{i}$ and $C_{j}$ is merged to $C_{k}$.
		(a) Before merging. (b)After merging, the centroid of new cluster $\mathbf{c}_{k}$ is different from both $\mathbf{c}_{i}$ and $\mathbf{c}_{j}$.		
	(c) The center node $\mathbf{x}_{l}$ is swapped from $C_{i}$ to $C_{j}$.
	Clustering before swapping: region \textcolor{blue}{\emph{Blue}} is $C_{i}$, and region \textcolor{green}{\emph{Green}} is $C_{j}$.
	\emph{Circle} represents nodes in clusters.
	(d) Clustering after swapping: region \textcolor{blue}{\emph{Blue}} is $C_{i'}$, and region \textcolor{green}{\emph{Green}} is $C_{j'}$.
	After the swapping operation, the belonging of node $\mathbf{x}_{l}$ is changed from $C_{i'}$ to $C_{j'}$.}
	\label{fig:merge}
\end{figure}

In the first step of the merging operation, each node is treated as an individual cluster, which forms potential merging pairs with its neighboring clusters. When a pair of clusters is merged to a new cluster, a merge cost is calculated and associated with this merge operation. For a merging operation $(C_{i}, C_{j}) \rightarrow C_{k}$, the merging cost is defined as: $E^{*}(C_{k})-E^{*}(C_{i})-E^{*}(C_{j})$. Fig. \ref{fig:merge} shows the concept of such an operation.

A heap is maintained to store all possible merging operations in the current clustering, paired with the corresponding costs as the key value. Next, the least-cost merging is performed. Each time after the least-cost pair is selected from the heap, only a local update is needed to maintain the validity of the merging heap: the remaining pairs of the two merged clusters in the heap are deleted, and the potential merging between the new cluster and its direct neighbors are inserted. This step is iteratively performed until the number of clusters reaches \emph{m}. As shown in Supplementary Material, the merging cost can be computed with $O(1)$ complexity, which is independent of the number of nodes in each cluster.

\textbf{Clustering Optimization by Swapping Operations:}
Only greedily merging the least-cost pair of clusters as initialization cannot guarantee the optimal solution for Eq. (\ref{eq:optimal-clustering}). The second stage of swapping operations can continue to optimize it based on the above initialization. In the greedy merging process, each time a pair of clusters is merged, nodes from both clusters are bound to reside in the same new cluster. Those nodes cannot freely decide where to go, so a swapping operation is necessary to relax the binding between nodes and clusters from the above initialization.

The swapping operation is defined as swapping a boundary node from its belonged cluster $C_{i}$ to swapping-available clusters. A boundary node $\mathbf{x}_{l}$ is the node which resides in $C_{i}$ and has at least a neighboring node $\mathbf{x}_{j} \in \mathcal{N}(\mathbf{x}_{l})$ that does not belong to $C_{i}$. We denote the set of clusters that $\mathcal{N}(\mathbf{x}_{l})$ reside in as swapping-available clusters ${NC}_{\mathbf{x}_{l}}$. Whether swapping $\mathbf{x}_{l}$ from $C_{i}$ to $C_{j} \in {NC}_{\mathbf{x}_{l}}$ is determined by the sign of energy change after the swapping operation. We call this energy change as swapping cost.

If the swapping cost is less than 0, it means this swapping can decrease the energy of our objective function Eq. (\ref{eqn:decompose}). Otherwise, the current clustering is best suitable for the tested node, and there is no further operation needed. If there is more than one cluster in ${NC}_{\mathbf{x}_{l}}$ that can optimize the clustering, we select the one that leads to the largest decrease of energy. To be more precise, as shown in the Supplementary Material, the swapping cost can be efficiently computed with $O(1)$ complexity, which is independent of the number of nodes in each cluster. Fig. \ref{fig:merge} (c) and (d) illustrates a typical swapping operation by swapping the center node $\mathbf{x}_{l}$ from $C_{i}$ to $C_{j}$ which results in new clusters $C_{i'}$ and $C_{j'}$.

In order to achieve real-time reconstruction, we need to accelerate the segmentation step. We only employ the merging operation after registering the mesh of canonical frame with the second frame. For the segmentation step of later frames, we initialize the clustering with the previous result and then perform swapping based on such initialization. For newly added nodes after depth fusion, their cluster belongings are determined by their closest existing neighbor nodes. Because of such initialization, the maintenance of heap structure is no longer needed. We can use GPU to compute the cross covariance matrix $\mathbf{A}^{t}(C_{n})$ and the energy $E^{*}(C_{n})$ in parallel according to Eqs. (\ref{eq:optimal-extend-energy}) and (\ref{eq:covariance}).  

\begin{figure}[h]
	\centering
	\includegraphics[width=0.9\textwidth]{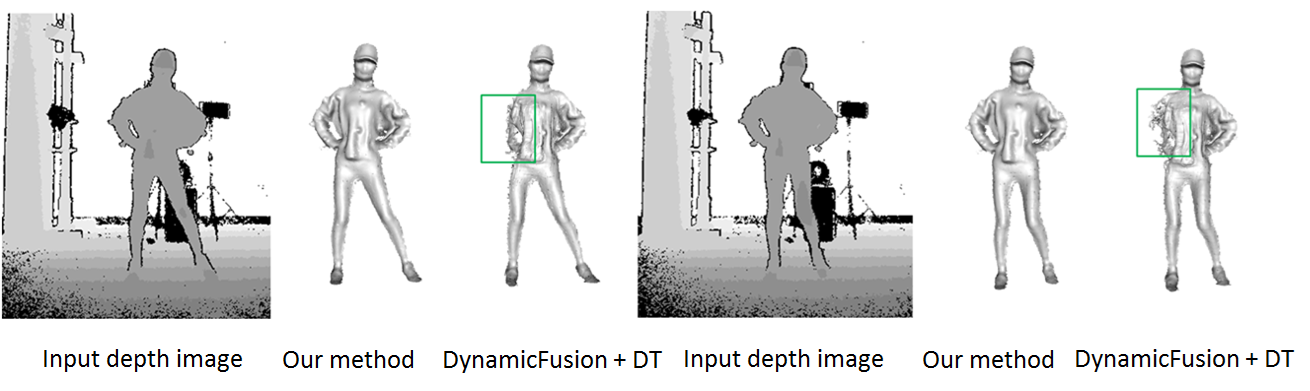}
	\caption{Segmentation improves the reconstruction result for fast inter-frame motion in direction parallel to the screen. In each group from left to right: input depth image, the reconstructed result of our method, and the result of DynamicFusion with only DT. }
	\label{fig:segmentImproveRes}
\end{figure}

Fig.~\ref{fig:segmentImproveRes} shows a comparison example between our method and DynamicFusion with DT in the registration step. Although both cases employ the DT-based correspondences computing, the reconstruction result of our method is much better because the introduction of segmentation.

The number of clusters can be given as a constant, or can be estimated dynamically by adding an energy threshold in the merging step. When the increased energy after one merging operation is bigger than the threshold, the merging step stops. This mechanism can automatically determine the number of clusters. Considering real-time performance, we can break any cluster with error higher than a given threshold into two new clusters and adjust the boundaries of new clusters in the swapping step. Cluster breaking can be achieved by merging all original nodes into two new clusters. Because only a small number of nodes in that cluster needs to be re-merged, the real-time performance can still hold.
Due to the space limit of the paper, details about dynamic clustering such as how the number of clusters influences the results, and the comparison of reconstruction results can be found in our Supplementary Material.
%-------------------------------------------------------------------------

\section{Results}
\begin{figure}[t]
	\centering
	\subfloat[Body turning]{\includegraphics[width=0.37\textwidth]{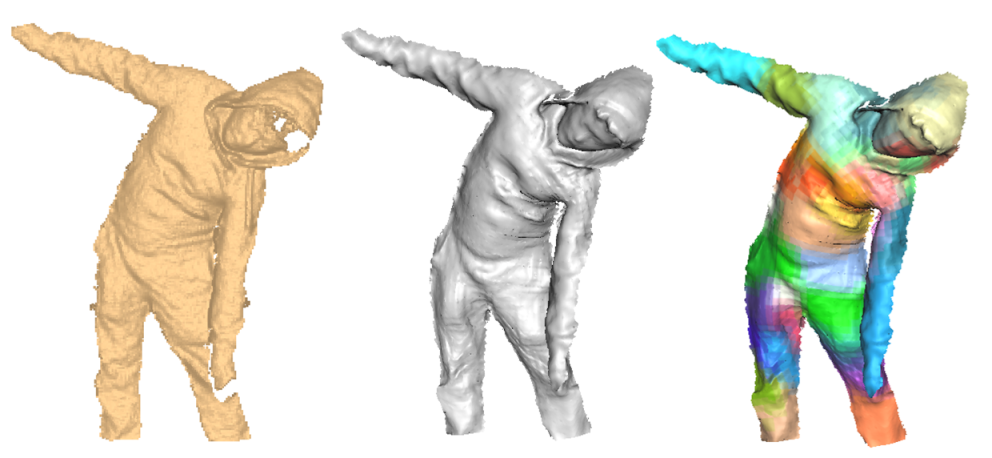}}
	~~\subfloat[Boxing]{\includegraphics[width=0.30\textwidth]{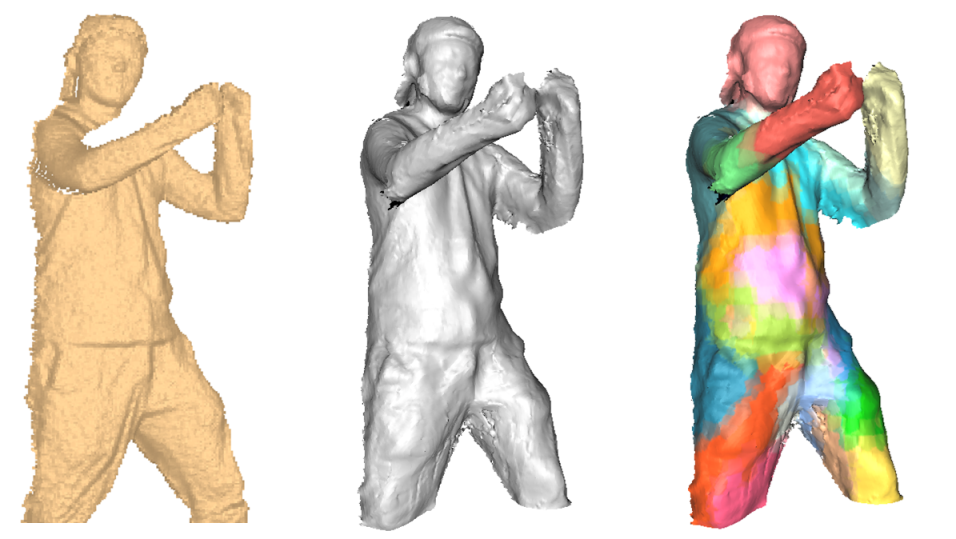}}
	~~\subfloat[Rolling arms]{\includegraphics[width=0.32\textwidth]{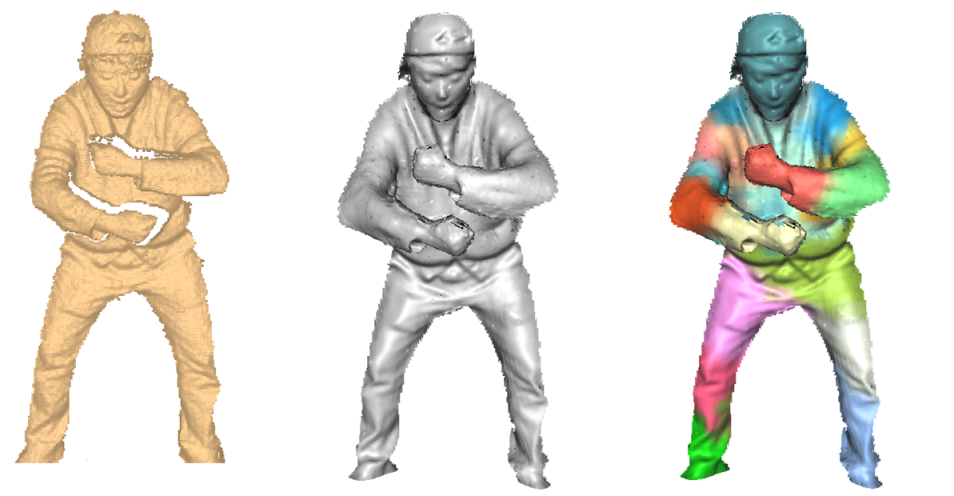}}
	\caption{Selected human motion reconstruction results by our system. From left to right for each motion: input depth, reconstructed geometry, segmentation.}
	\label{fig:ReconRes}
\end{figure}
\begin{figure}[t]
	\centering
	\subfloat[Bending cloth pipe]{\includegraphics[width=0.45\textwidth]{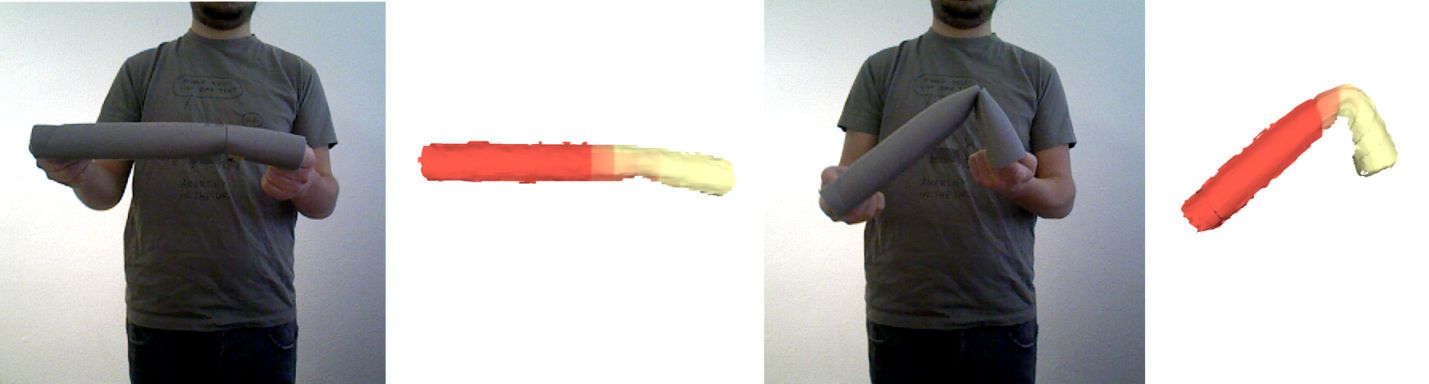}}
	\subfloat[Playing ``donkey'' hand puppet]{\includegraphics[width=0.45\textwidth]{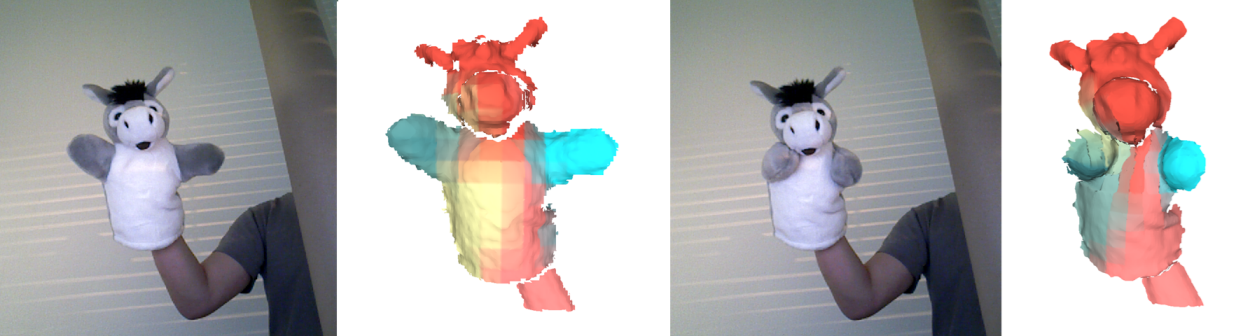}}
	\caption{Selected non-human reconstruction results by our system. (a) shows our reconstructed results of bending a cloth pipe at the 1/4 location; (b) shows our results of playing a ``donkey'' hand puppet.}
	\label{fig:nonHuman}
\end{figure}

In this section, we describe the performance of our system and details of its implementation, followed by qualitatively comparisons with state-of-the-art methods and evaluations. We captured more than 10 sequences with persons performing natural body motions like ``Boxing'', ``Dancing'', ``Body turning'', ``Rolling arms'', and ``Crossing arms'', etc. We have also experimented our algorithm on an existing dataset for articulated model reconstruction \cite{Tzionas:ECCVw:2016}.

Fig.~\ref{fig:ReconRes} shows some of our reconstruction results for motions ``Body turning'', ``Boxing'', and ``Rolling arms''. Our ArticulatedFusion system enables simultaneous geometry, motion, and segmentation reconstruction. As shown in Fig.~\ref{fig:ReconRes} (c), the human body is segmented by deformation clustering so hands, arms and head are segmented out because of their articulated motion property. 

Fig.~\ref{fig:nonHuman} shows that our system can also reconstruct geometry, motion, and segmentation for non-human motion sequences without any prior skeleton information or template. It automatically learns the segmentation from control nodes clustering. As shown in the 2nd and 4th columns of Fig.~\ref{fig:nonHuman} (a) and (b), faithful segmentation can be automatically generated during the reconstruction process with motions and fine geometry.

\subsection{Performance}
Our system is fully implemented on a single NVIDIA GeForce GTX 1080 graphics processing unit using both the OpenGL API and the NVIDIA CUDA API. The pipeline runs at 34--40 ms per frame on average. The time breaking of main steps is as follows (Table \ref{table:time}):  the preprocessing of the depth information (including bilateral filtering and calculation of the depth normals) requires 1 ms; the rendering of the results requires 1 ms. For two-level node motion optimization, we run 5 and 2 iterations respectively . In each iteration, to solve the linear equation, we run 10 iterations of PCG. The voxel resolution is 5 mm. For each vertex, 8 nearest nodes is used as its control node. The number of segments ranges from 6 to 40. In all examples, we capture the depth stream using a Kinect v2 with $512 \times 424$ depth image resolution. 

\begin{table*}[t]
	\centering
	\caption{Average computation time per frame for several motions (ms). Column ``Init'' is the time to initialize and update node graph. Column ``DT'' is the time to calculate distance transform. Columns ``Level 1'' and ``Level 2'' represent the time to solve level-1 and level-2 registration. Column ``TSDF'' represents the time to perform TSDF integration. Column ``Seg'' is the time of segmetation.}
	\label{table:time}
	\begin{tabular}{|l|l|l|l|l|l|l|l|l|l|}
		\hline
		& \begin{tabular}[c]{@{}l@{}}\# of \\ Segs\end{tabular} & \begin{tabular}[c]{@{}l@{}}\# of \\ Nodes\end{tabular} & \begin{tabular}[c]{@{}l@{}} Init\\(ms)\end{tabular} & \begin{tabular}[c]{@{}l@{}} DT\\(ms)\end{tabular} & 
		\begin{tabular}[c]{@{}l@{}}Level 1\\  (ms)\end{tabular} & 
		\begin{tabular}[c]{@{}l@{}}Level 2\\  (ms)\end{tabular} & 
		\begin{tabular}[c]{@{}l@{}}TSDF\\  (ms)\end{tabular} & \begin{tabular}[c]{@{}l@{}}Seg\\  (ms)\end{tabular} & \begin{tabular}[c]{@{}l@{}}Total\\  (ms)\end{tabular} \\ \hline
		Boxing & 20 &  1442&  2.7&  4.9&  8.3&  13.9&  4.7&  2.5& 37.0\\ \hline
		Rolling arms & 20 &  1914&  3.4&  4.6&  8.5&  15.0&  4.9&  2.7& 39.1\\ \hline
		Crossing arms & 12 &  1130&  2.5&  4.6&  7.1&  13.4&  5.1&  1.9& 34.6\\ \hline
		Dancing & 30 &  1569&  3.0&  4.7&  9.0&  14.4&  5.2&  3.0& 39.3\\ \hline
		Body turning & 20 &  2002&  3.5&  4.7&  8.6&  14.5&  4.8&  2.8& 38.9\\ \hline
	\end{tabular}
\end{table*}

\begin{figure}[t]
	\centering
	\subfloat[Input depth]{ 
		\label{subfig:armdepth} %% label for first subfigure 
		\includegraphics[width=0.41\textwidth]{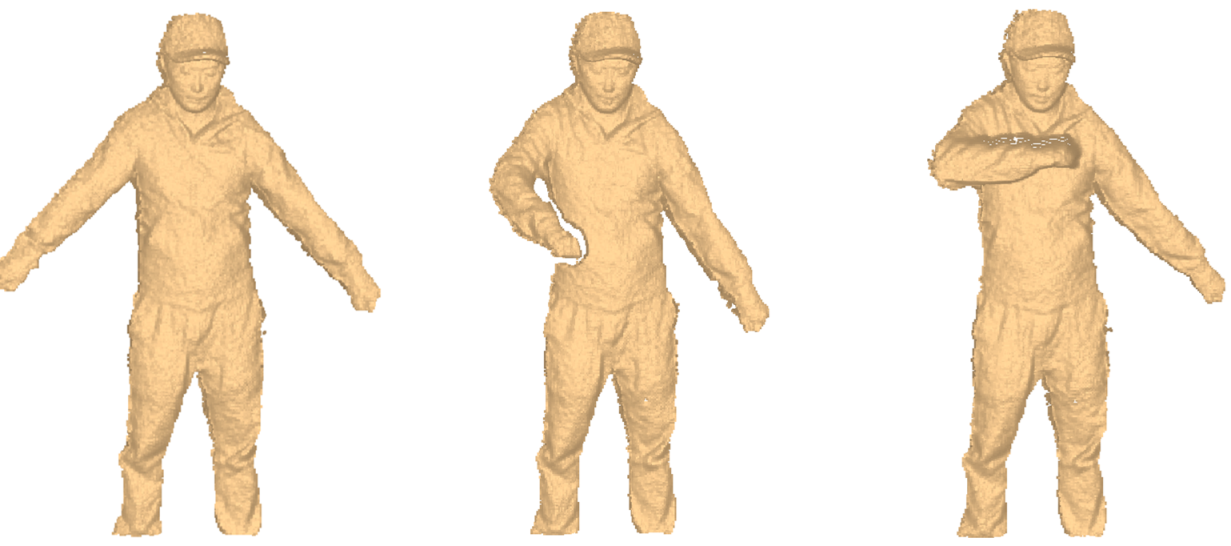}}
	\subfloat[Our method]{ 
		\label{subfig:arm1} %% label for first subfigure 
		\includegraphics[width=0.43\textwidth]{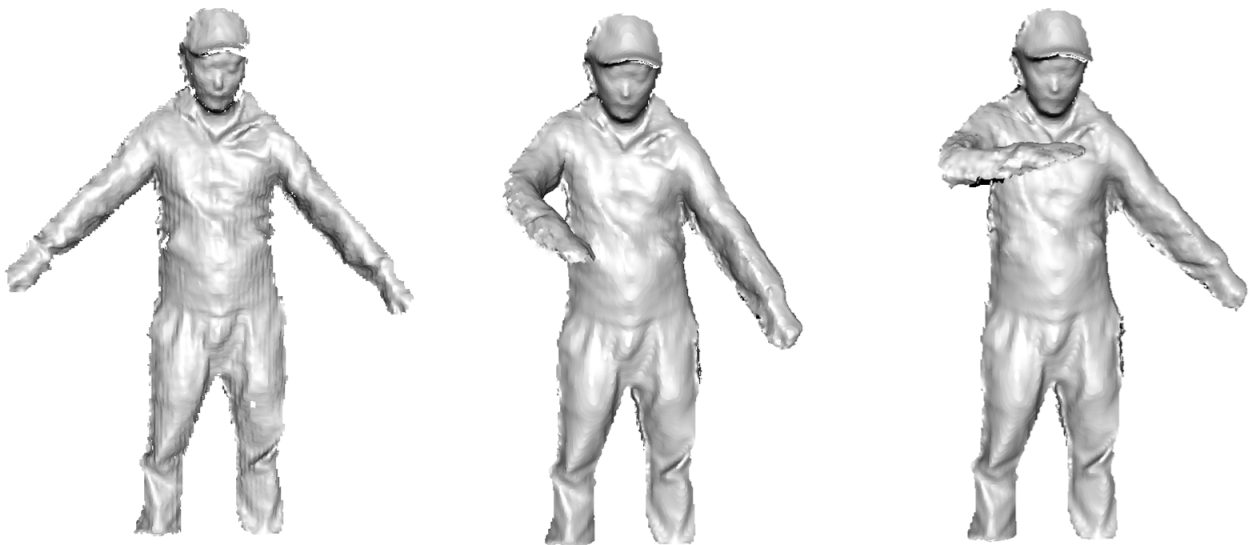}}\\
	\subfloat[DynamicFusion]{ 
		\label{subfig:arm2} %% label for first subfigure 
		\includegraphics[width=0.43\textwidth]{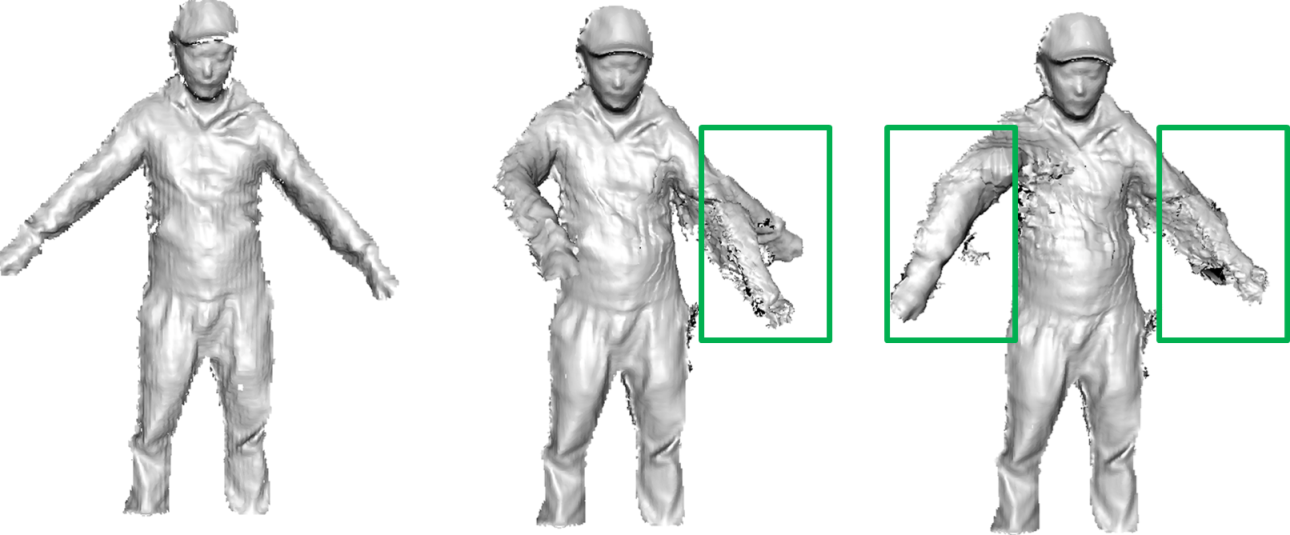}}
	\subfloat[VolumeDeform]{ 
		\label{subfig:arm3} %% label for first subfigure 
		\includegraphics[width=0.43\textwidth]{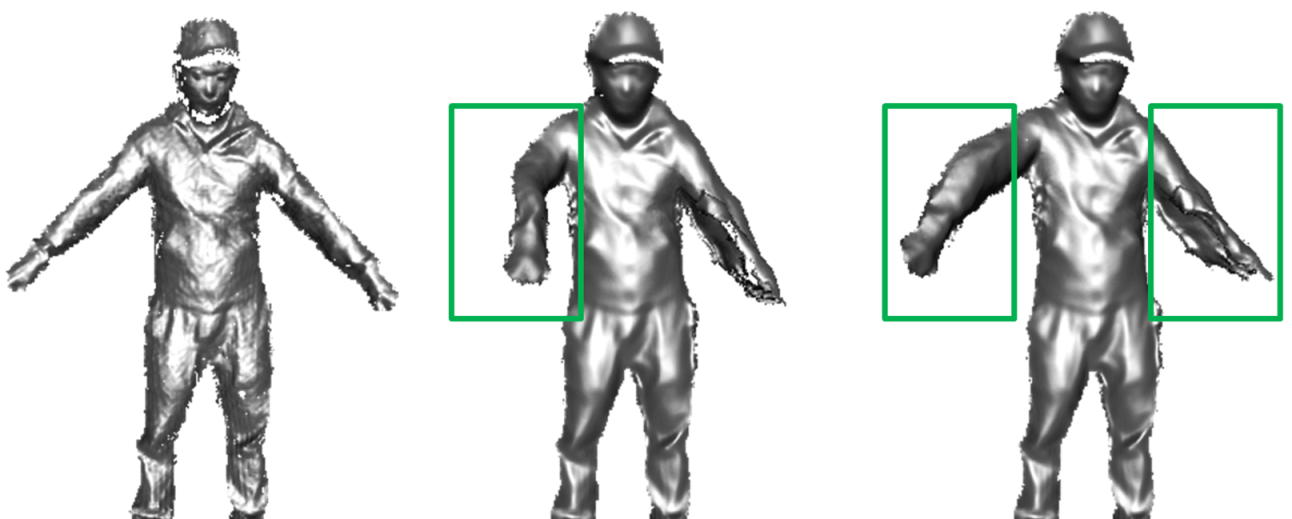}}
	
	\caption{Visual comparisons of the results between: (b) our method, (c) DynamicFusion~\cite{newcombe2015}, and (d) VolumeDeform~\cite{innmann2016volumedeform}, with input depth images shown in (a).}
	\label{fig:arm}
\end{figure}

\begin{figure}[t]
	\centering
	\includegraphics[width=1.0\textwidth]{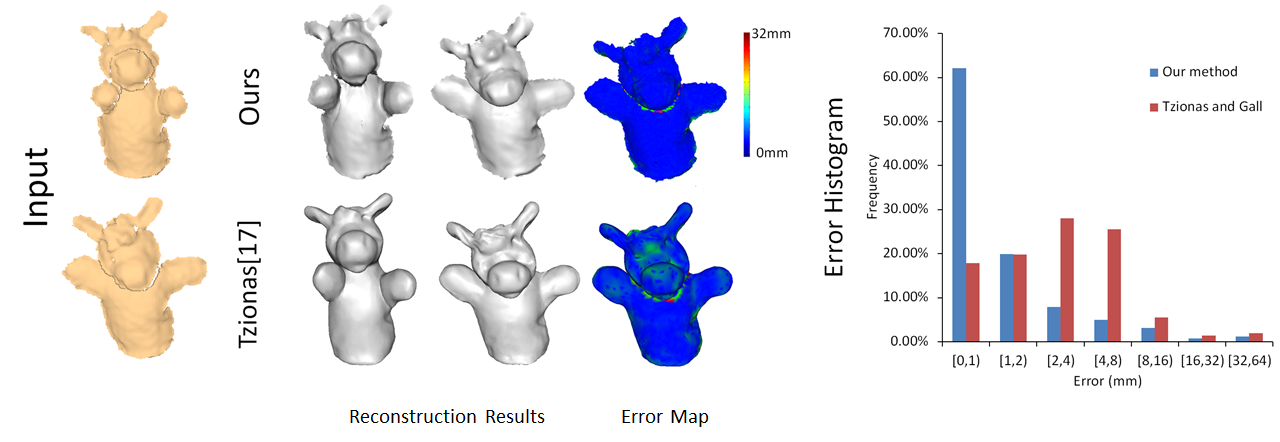}
	\caption{Non-human object reconstruction comparison on ``donkey'' hand puppet.}
	\label{fig:nonHumanCom}
\end{figure}

\subsection{Comparisons and Evaluations}
We compare our ArticulatedFusion with two state-of-the-art methods DynamicFusion~\cite{newcombe2015} and VolumeDeform~\cite{innmann2016volumedeform}.
Fig.~\ref{fig:arm} shows visual comparisons on motion ``Dancing''. We can see both DynamicFusion and VolumeDeform fail in the left and right arms region. Our method generates more faithful results for motions in tangential direction or motions having large occlusions. 

\begin{figure}[t]
	\centering
	\includegraphics[width=0.75\textwidth]{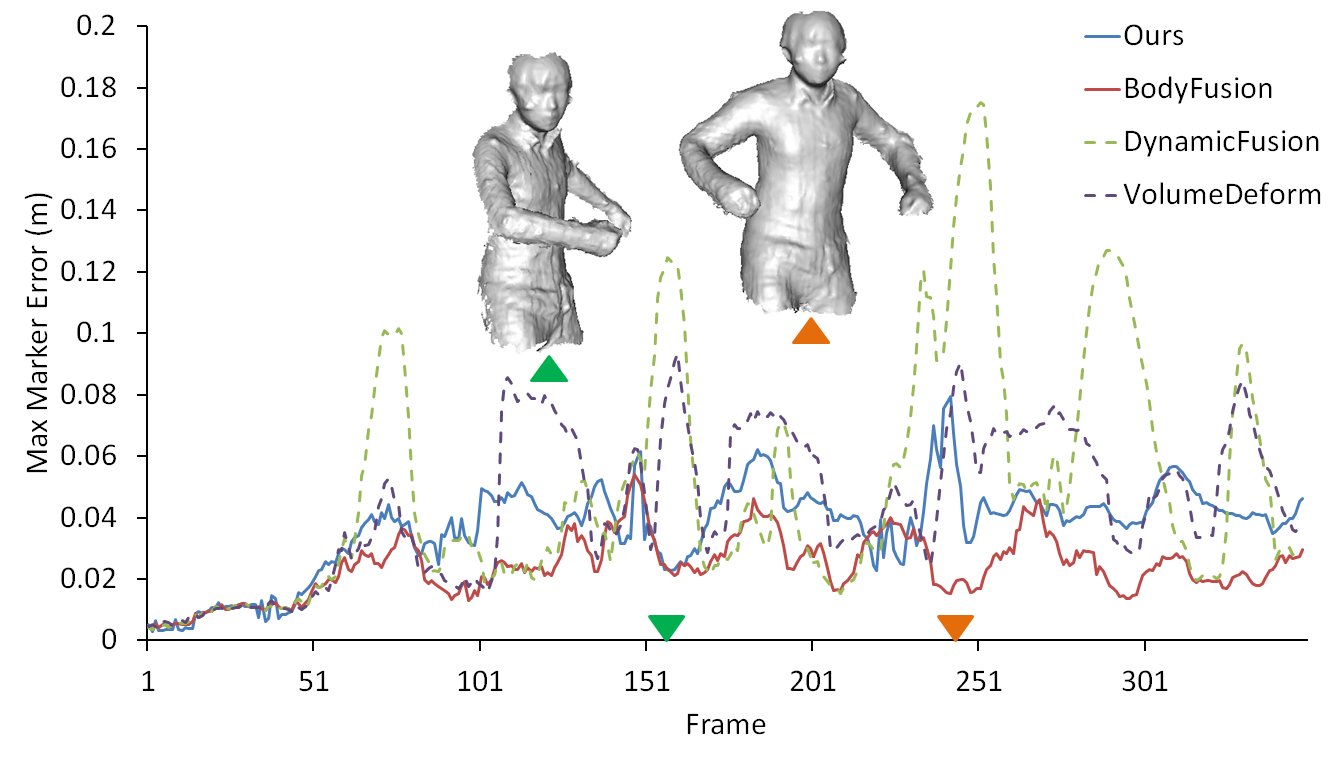}
	\caption{Quantitative comparison: max marker errors of our method, BodyFusion, DynamicFusion and VolumeDeform for a motion sequence.}
	\label{fig:bodyfusion}
\end{figure}

\begin{figure}[t]
		\centering
	\subfloat[]{ 
		\label{subfig:twoperson} %% label for first subfigure 
		\includegraphics[width=0.50\textwidth]{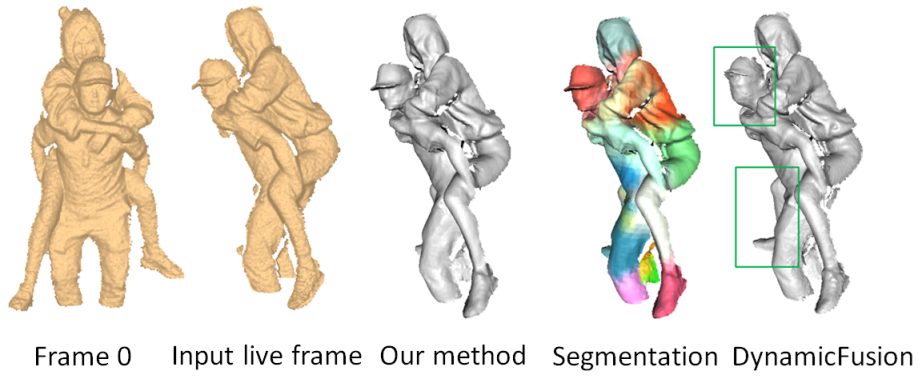}}
	\subfloat[]{ 
		\label{subfig:sunflower} %% label for first subfigure 
		\includegraphics[width=0.50\textwidth]{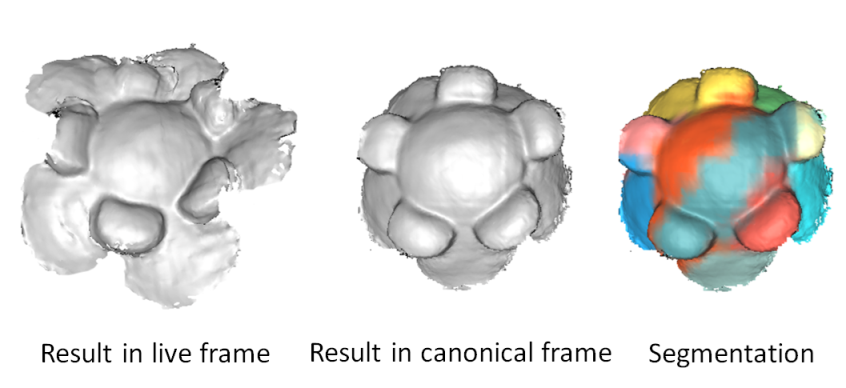}}
        
	\caption{
    (a) Reconstruction result comparison of our method and DynamicFusion~\cite{newcombe2015}.
    (b) Reconstruction result of the failure case in VolumeDeform~\cite{innmann2016volumedeform} (shown in their Fig. 9) for 5x speed input (skipping $5$ frames).}
	\label{fig:comparison}
\end{figure}

To further quantitatively evaluate our reconstructed segmentation and motion, we compare our results with the other state-of-the-art methods by using the Vicon-captured groundtruth data from BodyFusion~\cite{yu12bodyfusion}. 
In Fig.~\ref{fig:bodyfusion}, it is noted that our reconstruction error is comparative to BodyFusion (slightly higher though), but our method is more general and can be applied to dynamic scenes where Kinect-based skeleton is not available, such as non-human-body motions (Fig.~\ref{fig:nonHuman}, Fig.~\ref{fig:nonHumanCom}, and Fig.~\ref{fig:comparison} (b)) and human-body motions without initial skeleton information (Fig.~\ref{fig:comparison} (a)). In Fig.~\ref{fig:comparison} (a), the skeleton of the person on the back cannot be provided by Kinect because of high occlusion in the body. It is noted that the highlighted head and leg part is well reconstructed with the help of our segmentation, while they are not correctly tracked by DynamicFusion.

We compare our method with two other reconstruction methods that can reconstruct non-human objects.
Fig.~\ref{fig:nonHumanCom} shows a detailed comparison of the near-articulated example ``donkey'' hand puppet with the template-based reconstruction result in Tzionas and Gall's work~\cite{Tzionas:ECCVw:2016}. The first column of Fig.~\ref{fig:nonHumanCom} shows two input depth images. From both the error map and the error histogram, we can find our method has better error distribution than theirs. In order to have a fair comparison in error histogram, we only count visible vertices in both cases. Because of the introduction of segmentaion in the registration step, our method is more robust for fast motion. Fig.~\ref{fig:comparison} (b) shows another example of non-human object reconstruction. In VolumeDeform~\cite{innmann2016volumedeform}, their reconstruction fails when skipping 4 or more frames before next frame. But our method can still get a good result, while every petal of the sunflower is clustered as one segment. 

\section{Conclusion and Future Work} 
In this paper, we have seen that our two-level node optimization equipped efficient node graph segmentation enables better reconstructions for tangential and occluded motions, for non-rigid human and non-human motions captured with a single depth camera. We believe that our system represents a step forward towards a wider adoption of depth cameras in real-time applications, and opens the door for leveraging the high-level semantic information in reconstruction, e.g. differentiating dynamic and static scenes as shown in MixedFusion~\cite{zhang2017TVCG}.

Our system still has limitations in the reconstruction of very fast motions because of the blurred depth and our reliance on ICP-based local correspondence matching. Topological change of surfaces is also difficult to handle. In the future we would like to consider the integration of color information \cite{innmann2016volumedeform,guo2017} for further improvement on the motion optimization, and extracting a consistent tree-based skeleton structure from our segmentation.
\newline

\noindent \textbf{Acknowledgement} 
We would like to thank the reviewers for their valuable comments. We are grateful to Matthias Innmann for the help on comparison results of VolumeDeform, Tao Yu for providing their Vicon-based ground-truth marker data in BodyFusion, and Dimitrios Tzionas for providing their data. This work was partially supported by National Science Foundation under grant number IIS-1149737. Chao would like to thank the support provided by Hua Guo during the preparation for this paper.

%\noindent \textbf{Acknowledgement} 
%The authors would like to thank the reviewers for their valuable comments. Chao also would like to thank the support provided by Hua Guo during the preparation for this paper and Zheheng Zhao for his assistance in video editing.

\bibliographystyle{splncs}
\bibliography{egbib}
\clearpage

\section*{\textemdash Supplementary Material\textemdash}
\section*{Formula Derivation}
In this section, we first introduce all symbols and then the derivation of related formula. The list of all symbols is shown in table~\ref{table:symbol}:
\begin{table}
	\centering
	\caption{List of all symbols}
	\label{table:symbol}
	\begin{tabular}[width=\textwidth]{|l|l|}	
		\hline
		$C_{k}$& the $k$-th cluster \\ \hline
		$\mathbf{x}$& the position in canonical frame for a node of cluster $C_{k}$ \\ \hline
		$\mathbf{y}^{t}$& the position in live frame $t$ for a node in cluster $C_{k}$  \\ \hline
		$\mathbf{c}_{k}$& the centroid position of cluster $C_{k}$ in canonical frame \\ \hline
		$\mathbf{c}^{t}_{k}$& the centroid position of cluster $C_{k}$ in live frame $t$ \\ \hline
		$n_{k}$& the number of nodes belonging to cluster $C_{k}$ in canonical frame \\ \hline
		$\mathbf{A}^{t}(C_{k})$& the cross covariance matrix of cluster $C_{k}$ \\ \hline
		$(R^{t}_{k}, \mathbf{t}^{t}_{k})$& rotation and translation of cluster $C_{k}$ \\ \hline
		$(R^{*}, \mathbf{t}^{*})$& optimal rotation and translation of cluster $C_{k}$\\ \hline
		$\sigma_{kq}$& the $q$-th singular value of $\mathbf{A}^{t}(C_{k})$\\ \hline
	\end{tabular}
\end{table}

%--------------------
\noindent \textbf{Optimal Energy $E^{*}$ for an Arbitary Cluster}
\newline
The total segmentation energy is as follows:
\begin{equation}
\label{seg-energy}
E_{seg} = \sum_{k=1}^{m}\sum_{\mathbf{x}\in C_{k}}\|R_{k}^{t}\mathbf{x}+\mathbf{t}_{k}^{t}-\mathbf{y}^{t}\|^{2}.
\end{equation}
When segmentation is fixed, the segmentation energy term of each cluster is independent to each other. Therefore, we can parallelly compute the optimal energy of each cluster $E^{*}(C_{k})$.
In order to calculate $E^{*}(C_{k})$, we first need to know how to compute the optimal value of rotation and translation $\left(R_{n}^{t}, \mathbf{t}_{n}^{t}\right)$ based on the segmentation energy term of an arbitary cluster $C_{k}$:
\begin{equation}
\label{eq:one-cluster-energy}
E(C_{k}) =  \sum_{\mathbf{x}\in C_{k}}\|R_{k}^{t}\mathbf{x}+\mathbf{t}_{k}^{t}-\mathbf{y}^{t}\|^{2}.
\end{equation}

According to Sorkine-Hornung and Rabinovich's technical report~\cite{sorkine2016least}, the optimal $\left(R_{n}^{t}, \mathbf{t}_{n}^{t}\right)$ is as follows:
\begin{equation}
\label{R}
R^{*}=V\begin{pmatrix}
1&  &  \\ 
&  1&  \\ 
&  &  det(VU^{\top})\\ 
\end{pmatrix}U^{\top},
\end{equation}

\begin{equation}
\label{t}
\mathbf{t}^{*} = \mathbf{c}_{k}^{t} - R^{*}\mathbf{c}_{k} ,
\end{equation}
where 
\begin{equation}
\label{S}
\mathbf{A}^{t}(C_{k}) = PQ^{\top} = U \Sigma V^{\top},
\end{equation}

\begin{equation}
\label{P}
P = \begin{pmatrix}
...  \\ 
\mathbf{x}-\mathbf{c}_{k}  \\ 
...\\ 
\end{pmatrix},
\end{equation}

\begin{equation}
\label{Q}
Q = \begin{pmatrix}
...  \\ 
\mathbf{y}^{t}-\mathbf{c}^{t}_{k}  \\ 
...\\ 
\end{pmatrix},
\end{equation}

\begin{equation}
\label{x-y}
PQ^{\top} =\sum_{\mathbf{x} \in C_{k}}(\mathbf{x}-\mathbf{c}_{k})(\mathbf{y}^{t}-\mathbf{c}^{t}_{k})^{\top},
\end{equation}

\begin{equation}
\label{c}
\mathbf{c}_{k} = \frac{\sum_{\mathbf{x}\in C_{k}}\mathbf{x}}{n_{k}},
\end{equation}

\begin{equation}
\label{c_t}
\mathbf{c}_{k}^{t} = \frac{\sum_{\mathbf{y}^{t}\in C_{k}}\mathbf{y}^{t}}{n_{k}}.
\end{equation}

$U \Sigma V^{\top}$ is the Singular Value Decomposition (SVD) of matrix $\mathbf{A}^{t}(C_{k})$.

Replace $\mathbf{t}_{k}^{t}$ in Eq.(\ref{eq:one-cluster-energy}) by $\mathbf{t}^{*}$ in Eq.(\ref{t}), we have:
\begin{equation}
\label{eq:optimal-energy}
\begin{split}
E^{*}(C_{k}) = & \sum_{\mathbf{x}\in C_{k}}\|R^{*}\mathbf{x}+\mathbf{t}^{*}-\mathbf{y}^{t}\|^{2} \\
 = & \sum_{\mathbf{x}\in C_{k}}\|R^{*}(\mathbf{x}-\mathbf{c}_{k})-(\mathbf{y}^{t}-\mathbf{c}_{k}^{t})\|^{2} \\
 = & \sum_{\mathbf{x} \in C_{k}}[(\mathbf{x}-\mathbf{c}_{k})^{\top}(\mathbf{x}-\mathbf{c}_{k})
+(\mathbf{y}^{t}-\mathbf{c}^{t}_{k})^{\top}(\mathbf{y}^{t}-\mathbf{c}^{t}_{k})] \\
 & - 2\sum_{\mathbf{x} \in C_{k}}(\mathbf{y}^{t}-\mathbf{c}^{t}_{k})^{\top}R^{*}(\mathbf{x}-\mathbf{c}_{k}).
\end{split}
\end{equation}

By making $\mathbf{p} := \mathbf{x}-\mathbf{c}_{k}$ and $\mathbf{q} := \mathbf{y}^{t}-\mathbf{c}^{t}_{k}$, according to Sorkine-Hornung and Rabinovich's technical report \cite{sorkine2016least},
we have:
\begin{equation}
\label{eq:trace}
\begin{split}
&\sum_{\mathbf{x} \in C_{k}}(\mathbf{y}^{t}-\mathbf{c}^{t}_{k})^{\top}R^{*}(\mathbf{x}-\mathbf{c}_{k})\\
&=\sum_{\mathbf{x} \in C_{k}}\mathbf{q}^{\top}R^{*}\mathbf{p}\\
&=tr(Q^{\top}R^{*}P) = tr(R^{*}PQ^{\top}) = tr(R^{*}U\Sigma V^{\top})\\
& = tr(\Sigma V^{\top}R^{*}U).
\end{split}
\end{equation}

According to Eq.(\ref{R}), Eq.(\ref{eq:trace}) can be re-written as:
\begin{equation}
\label{eq:singular}
\sum_{\mathbf{x} \in C_{k}}(\mathbf{y}^{t}-\mathbf{c}^{t}_{k})^{\top}R^{*}(\mathbf{x}-\mathbf{c}_{k})=\sum_{q=1}^{3}\sigma_{kq}.
\end{equation}

Therefore,
\begin{equation}
\label{eq:merge-energy-total}
E^{*}(C_{k})
=\sum_{\mathbf{x} \in C_{k}}[(\mathbf{x}-\mathbf{c}_{k})^{\top}(\mathbf{x}-\mathbf{c}_{k})
+(\mathbf{y}^{t}-\mathbf{c}^{t}_{k})^{\top}(\mathbf{y}^{t}-\mathbf{c}^{t}_{k})]-2\sum_{q=1}^{3}\sigma_{kq}.
\end{equation}

%--------------------
\noindent \textbf{Cluster Mergence: $(C_{i}, C_{j})\rightarrow C_{k}$}
\newline
By decomposing Eq.(\ref{eq:merge-energy-total}) into two parts, we have
\begin{equation}
\label{eq:merge-energy1}
E_{1}^{*}(C_{k})=\sum_{\mathbf{x} \in C_{k}}[(\mathbf{x}-\mathbf{c}_{k})^{\top}(\mathbf{x}-\mathbf{c}_{k})
+(\mathbf{y}^{t}-\mathbf{c}^{t}_{k})^{\top}(\mathbf{y}^{t}-\mathbf{c}^{t}_{k})],
\end{equation}

\begin{equation}
\label{eq:merge-energy2}
E_{2}^{*}(C_{k})=-2\sum_{q=1}^{3}\sigma_{kq}.
\end{equation}

If we merge a pair of clusters $(C_{i}, C_{j})$ into a new cluster $C_{k}$, $E_{1}^{*}(C_{k})$ and $E_{2}^{*}(C_{k})$ can be updated in constant time.

Firstly,
\begin{equation}
\label{eq:merge-covariance}
\begin{split}
\mathbf{A}^{t}(C_{k}) = & \sum_{\mathbf{x} \in C_{k}}(\mathbf{x}-\mathbf{c}_{k})(\mathbf{y}^{t}-\mathbf{c}^{t}_{k})^{\top}\\
= & \sum_{\mathbf{x} \in C_{i}}(\mathbf{x}-\mathbf{c}_{k})(\mathbf{y}^{t}-\mathbf{c}^{t}_{k})^{\top}+\sum_{\mathbf{x} \in C_{j}}(\mathbf{x}-\mathbf{c}_{k})(\mathbf{y}^{t}-\mathbf{c}^{t}_{k})^{\top}\\
= & n_{i}(\mathbf{c}_{i}-\mathbf{c}_{k})(\mathbf{c}^{t}_{i}-\mathbf{c}^{t}_{k})^{\top}
+n_{j}(\mathbf{c}_{j}-\mathbf{c}_{k})(\mathbf{c}^{t}_{j}-\mathbf{c}^{t}_{k})^{\top}\\
& +\mathbf{A}^{t}(C_{i})+\mathbf{A}^{t}(C_{j}),
\end{split}
\end{equation}
where $n_{i}$ is the number of nodes in cluster $C_{i}$, and so is $n_{j}$ for $C_{j}$.
The new centroid $\mathbf{c}_{k}$ of cluster $C_{k}$ can be updated by:
\begin{equation}
\label{eq:centroid-k}
\mathbf{c}_{k}=\frac{n_{i}\mathbf{c}_{i}+n_{j}\mathbf{c}_{j}}{n_{i}+n_{j}}.
\end{equation}
So is $\mathbf{c}^{t}_{k}$. And $\sigma_{kq}$ can be solved by the SVD decomposition of $A^{t}(C_{k})$.
Therefore, $E_{2}^{*}(C_{k})$ can be updated in constant time.

Secondly,
\begin{equation}
\label{eq:merge-energy-sum}
\begin{split}
E_{1}^{*}(C_{k}) = & E_{1}^{*}(C_{i})+E_{1}^{*}(C_{j})\\
& +n_{i}(\mathbf{c}_{i}-\mathbf{c}_{k})^{\top}(\mathbf{c}_{i}-\mathbf{c}_{k})
+n_{i}(\mathbf{c}^{t}_{i}-\mathbf{c}^{t}_{k})^{\top}(\mathbf{c}^{t}_{i}-\mathbf{c}^{t}_{k})\\
& +n_{j}(\mathbf{c}_{j}-\mathbf{c}_{k})^{\top}(\mathbf{c}_{j}-\mathbf{c}_{k})
+n_{j}(\mathbf{c}^{t}_{j}-\mathbf{c}^{t}_{k})^{\top}(\mathbf{c}^{t}_{j}-\mathbf{c}^{t}_{k}).
\end{split}
\end{equation}
Therefore, $E_{1}^{*}(C_{k})$ can be updated in constant time.

Because both $E_{1}^{*}(C_{k})$ and $E_{2}^{*}(C_{k})$ can be updated in constant time, the energy change aftering merging $C_{i}$, $C_{j}$ into $C_{k}$ can be calculated in constant time as follows: 
\begin{equation}\label{eq:delta-energy1}
\begin{split}
\Delta E_{1}^{*} = & E_{1}^{*}(C_{k})-E_{1}^{*}(C_{i})-E_{1}^{*}(C_{j})\\
= & n_{i}(\mathbf{c}_{i}-\mathbf{c}_{k})^{\top}(\mathbf{c}_{i}-\mathbf{c}_{k})
+n_{i}(\mathbf{c}^{t}_{i}-\mathbf{c}^{t}_{k})^{\top}(\mathbf{c}^{t}_{i}-\mathbf{c}^{t}_{k})\\
& +n_{j}(\mathbf{c}_{j}-\mathbf{c}_{k})^{\top}(\mathbf{c}_{j}-\mathbf{c}_{k})
+n_{j}(\mathbf{c}^{t}_{j}-\mathbf{c}^{t}_{k})^{\top}(\mathbf{c}^{t}_{j}-\mathbf{c}^{t}_{k}),
\end{split}
\end{equation}

\begin{equation}\label{eq:delta-energy2}
\Delta E_{2}^{*}=E_{2}^{*}(C_{k})-E_{2}^{*}(C_{i})-E_{2}^{*}(C_{j}),
\end{equation}

\begin{equation}\label{eq:delta-merge-energy}
\Delta E^{*}=\Delta E_{1}^{*}+\Delta E_{2}^{*}.\\
\end{equation}

In summary, the merge cost $\Delta E^{*}$ can be computed efficiently by keeping track of the centroid, the number of nodes and the cross covariance matrix of each cluster: $\{\mathbf{c}_{k}, n_{k}, \mathbf{A}^{t}(C_{k})\}$, simple computing operations in constant time and a SVD decomposition of a $3\times3$ cross covariance matrix.

\noindent \textbf{Cluster Optimization: Swapping $C_{l}$ from $C_{i}$ to $C_{j}$}
\newline
Let us suppose swapping a one-node cluster $C_{l}$ from $C_{i}$ to $C_{j}$ ($C_{l}$ only contains one boundary node $\mathbf{x}_{l}$).
After swapping, $C_{i}$ becomes $C_{i'}$ and $C_{j}$ becomes $C_{j'}$: i.e., $C_{i'}=C_{i} - C_{l}$, $C_{j'}=C_{j}\cup C_{l}$.
Then we can consider that $C_{i}$ is formed by a merging operation $(C_{i'}, C_{l})\rightarrow C_{i}$ and $C_{j'}$ is formed by a merging operation $(C_{j}, C_{l})\rightarrow C_{j'}$.
This can help us treat the swapping operation as a combination of merging operations, and calculate energy change by using equations from the previous section.

The decrease of energy $E_{2}^{*}$ is:
\begin{equation}\label{eq:swap-delta-energy2}
\Delta E_{2}^{*}=E_{2}^{*}(C_{i'})+E_{2}^{*}(C_{j'})-E_{2}^{*}(C_{i})-E_{2}^{*}(C_{j}).
\end{equation}
$E_{2}^{*}(C_{i'})+E_{2}^{*}(C_{j'})$ can be calculated by SVD of $\mathbf{A}^{t}(C_{i'})$ and $\mathbf{A}^{t}(C_{j'})$.
$\mathbf{A}^{t}(C_{i'})$ and $\mathbf{A}^{t}(C_{j'})$ can be updated by proper substitution according to Eq. (\ref{eq:merge-covariance}):
\begin{equation}
\label{eq:swap-covariance-i}
\begin{split}
\mathbf{A}^{t}(C_{i'}) = & \mathbf{A}^{t}(C_{i})-\mathbf{A}^{t}(C_{l})\\
& -n_{i'}(\mathbf{c}_{i'}-\mathbf{c}_{i})(\mathbf{c}^{t}_{i'}-\mathbf{c}^{t}_{i})^{\top}
-(\mathbf{x}_{l}-\mathbf{c}_{i})(\mathbf{y}^{t}_{l}-\mathbf{c}^{t}_{i})^{\top},
\end{split}
\end{equation}
\begin{equation}
\label{eq:swap-covariance-j}
\begin{split}
\mathbf{A}^{t}(C_{j'}) = & \mathbf{A}^{t}(C_{j})+\mathbf{A}^{t}(C_{l})\\
& +n_{j}(\mathbf{c}_{j}-\mathbf{c}_{j'})(\mathbf{c}^{t}_{j}-\mathbf{c}^{t}_{j'})^{\top}
+(\mathbf{x}_{l}-\mathbf{c}_{j'})(\mathbf{y}^{t}_{l}-\mathbf{c}^{t}_{j'})^{\top}.
\end{split}
\end{equation}

Then following the same rules in calculation of $\Delta E_{2}^{*}$, the decrease of energy $E_{1}^{*}$ is:
\begin{equation}\label{eq:swap-delta-energy1}
\begin{split}
\Delta E_{1}^{*} = & E_{1}^{*}(C_{i'})+E_{1}^{*}(C_{j'})-E_{1}^{*}(C_{i})-E_{1}^{*}(C_{j})\\
= & n_{j}(\mathbf{c}_{j}-\mathbf{c}_{j'})^{\top}(\mathbf{c}_{j}-\mathbf{c}_{j'})
+n_{j}(\mathbf{c}^{t}_{j}-\mathbf{c}^{t}_{j'})^{\top}(\mathbf{c}^{t}_{j}-\mathbf{c}^{t}_{j'})\\
& +(\mathbf{x}_{l}-\mathbf{c}_{j'})^{\top}(\mathbf{x}_{l}-\mathbf{c}_{j'})+(\mathbf{y}^{}_{l}-\mathbf{c}^{}_{j'})^{\top}(\mathbf{y}^{t}_{l}-\mathbf{c}^{t}_{j'})\\
& -n_{i'}(\mathbf{c}_{i'}-\mathbf{c}_{i})^{\top}(\mathbf{c}_{i'}-\mathbf{c}_{i})
-n_{i'}(\mathbf{c}^{t}_{i'}-\mathbf{c}^{t}_{i})^{\top}(\mathbf{c}^{t}_{i'}-\mathbf{c}^{t}_{i})\\
& -(\mathbf{x}_{l}-\mathbf{c}_{i})^{\top}(\mathbf{x}_{l}-\mathbf{c}_{i})-(\mathbf{y}^{t}_{l}-\mathbf{c}^{t}_{i})^{\top}(\mathbf{y}^{t}_{l}-\mathbf{c}^{t}_{i}).
\end{split}
\end{equation}

In summary, the swapping cost can be computed efficiently by keeping track of the centroid, the number of nodes, and the cross covariance matrix of each cluster: $\{\mathbf{c}_{k}, n_{k}, \mathbf{A}^{t}(C_{k})\}$, simple computing operations in costant time and SVD decompositions of two $3\times3$ cross covariance matrices.

\section*{Discussion on Dynamic Clustering}

\begin{figure}[]
	\centering
	\subfloat[]{\includegraphics[width=0.33\textwidth]{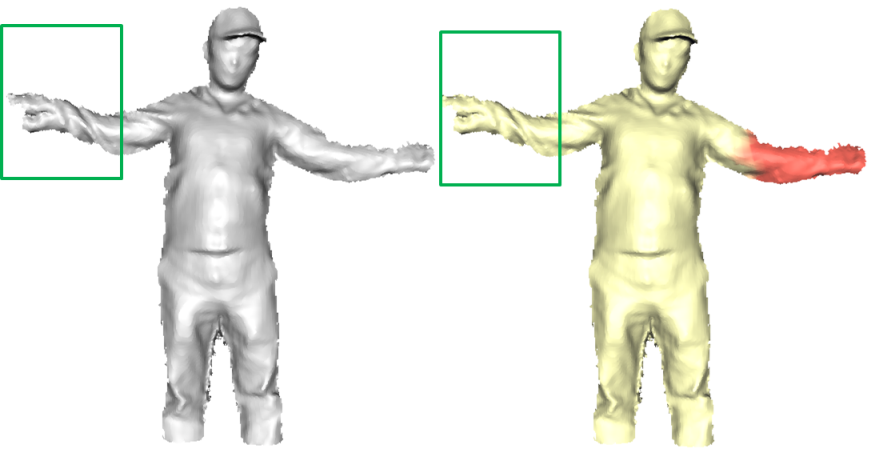}}
	~\subfloat[]{\includegraphics[width=0.33\textwidth]{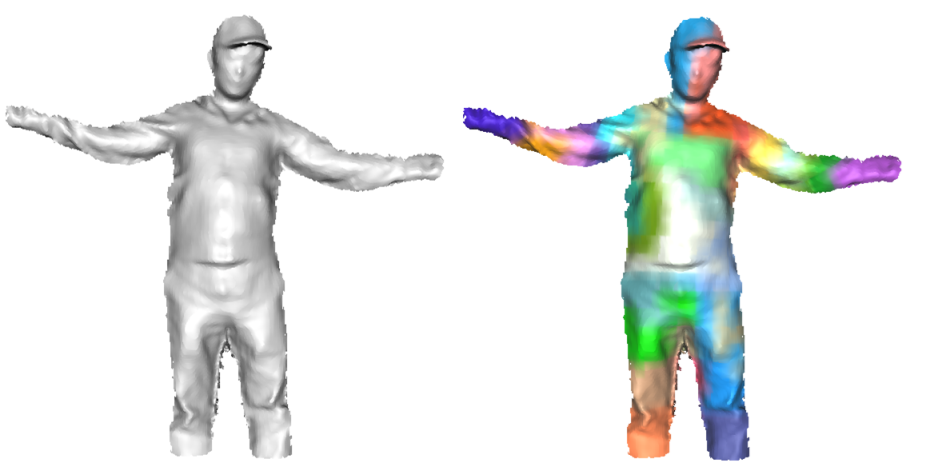}}	
	~\subfloat[]{\includegraphics[width=0.33\textwidth]{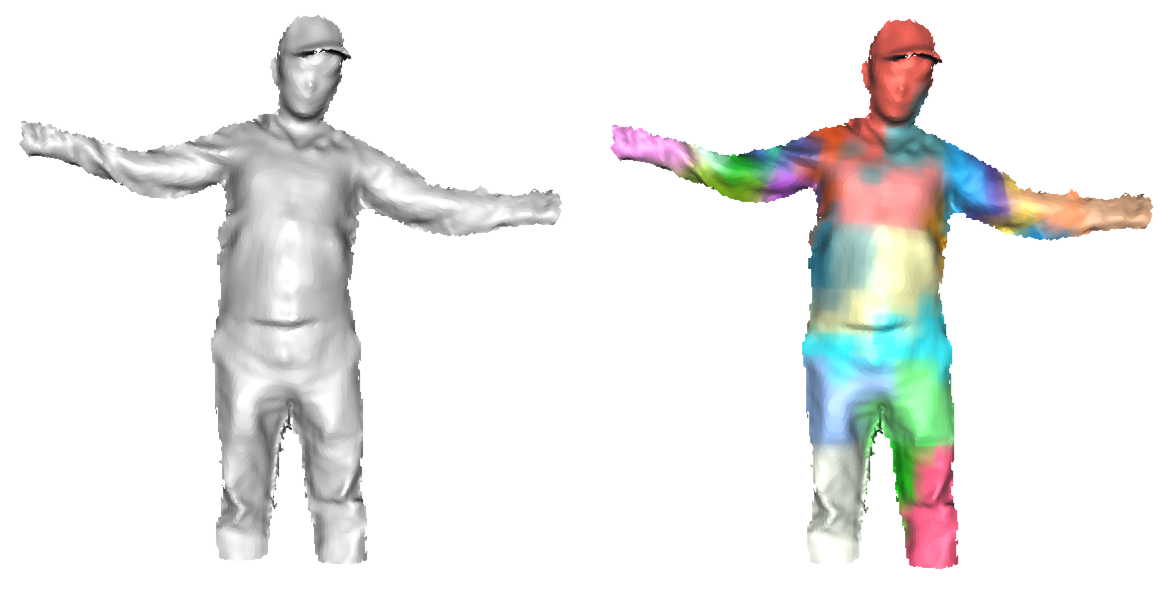}}
	
	\caption{(a) Reconstruction result with fixed 2 clusters.
		     (b) Reconstruction result with fixed 30 clusters.
	         (c) Reconstruction result with dynamic clustering.
	         For each group of subfigures: the left-hand side image is the reconstructed geometry and the right-hand side image is current segmentation.}
	\label{fig:cluster}
\end{figure}

\noindent \textbf{Influence of Different Number of Clusters on Results}
\newline
Suppose $n$ is the necessary number of clusters for a motion, $m$ is the total number of nodes, and $k$ is the fixed number of clusters to choose. There are 3 situations:
\begin{enumerate}
	\item $k<n$: i.e., insufficient number of clusters. The reconstruction result will be bad (Fig.~\ref{fig:cluster} (a)). The highlighted right arm part is not well reconstructed.
	\item $k\geq n$ and $k\ll m$: the reconstruction result will be good (Fig.~\ref{fig:cluster} (b) and (c)). 
	\item Otherwise, the situation becomes similar as ``DT + DynamicFusion''. The reconstruction result will be still bad. 
\end{enumerate}

\vspace{-1.5em}

\begin{figure}[]
	\centering
	\subfloat[]{\includegraphics[width=0.33\textwidth]{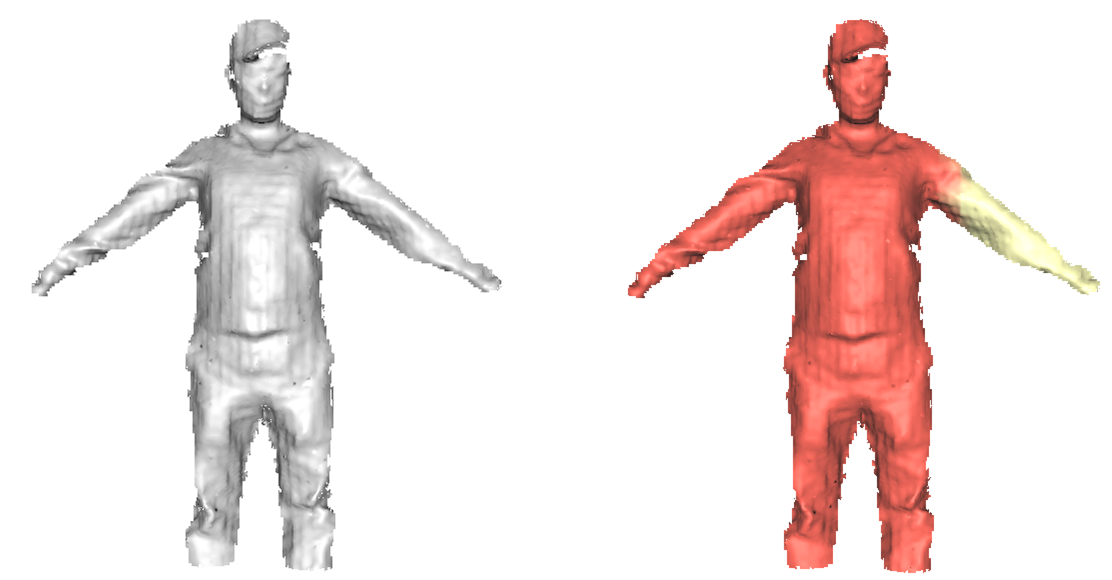}}
	~\subfloat[]{\includegraphics[width=0.33\textwidth]{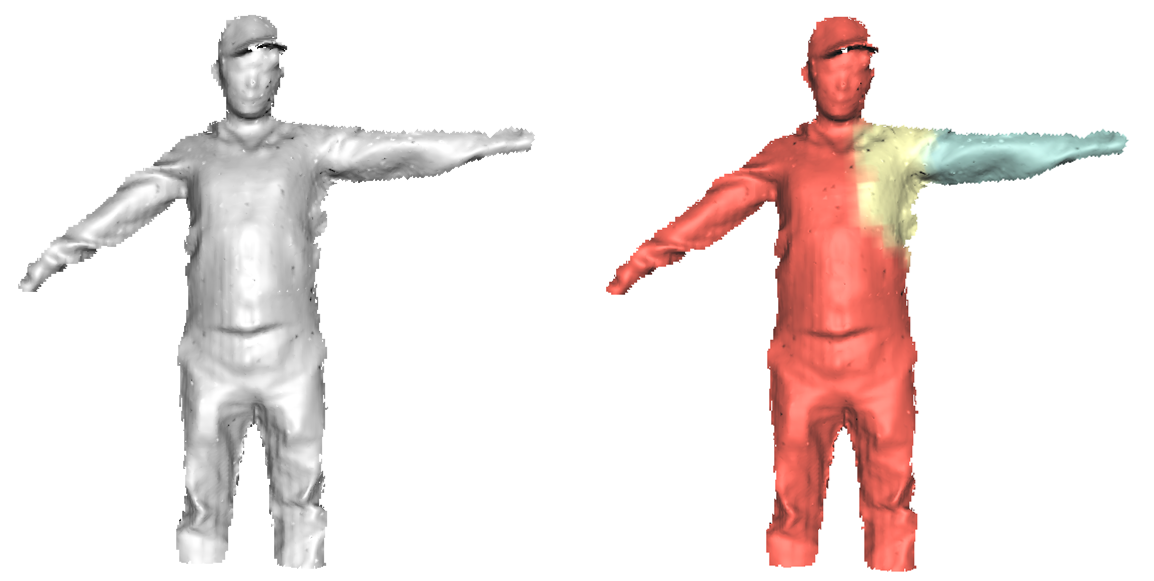}}	
	~\subfloat[]{\includegraphics[width=0.33\textwidth]{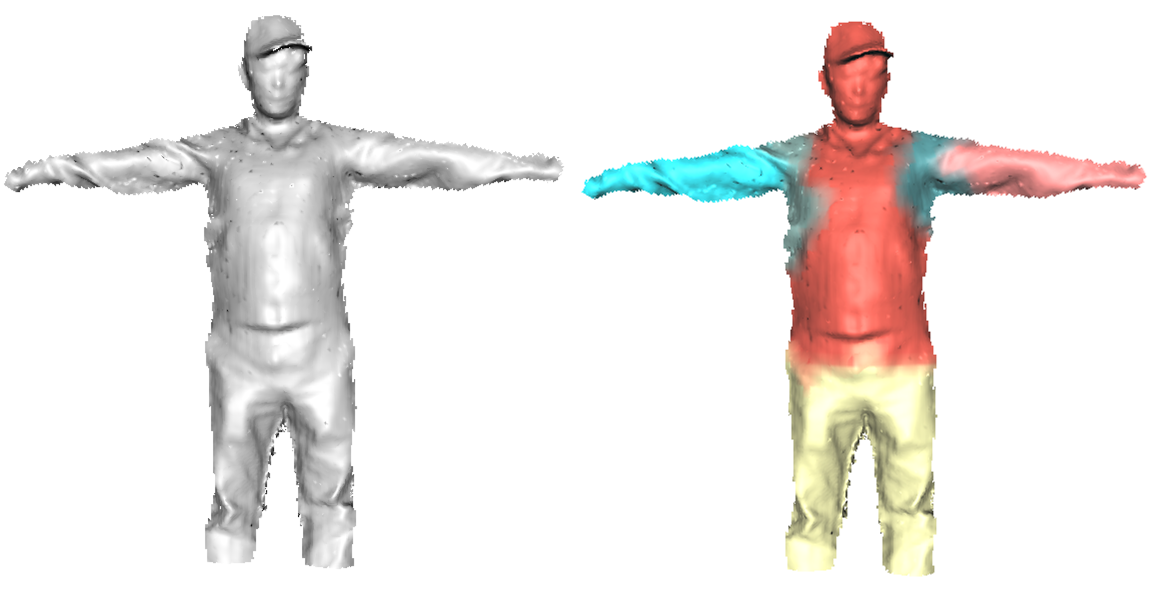}}
	
	\caption{Dynamic Clustering mechanism. 
    From (a) to (b): number of clusters is changed from 2 to 3 when one arm is raising.
	From (b) to (c): number of clusters is changed from 3 to 6 when two arms are raising.
	For each group of subfigures: the left-hand side image is the reconstructed geometry and the right-hand side image is current segmentation.}
	\label{fig:dynamicCluster}
\end{figure}

Dynamic clustering mechanism will automatically determine the number of clusters by setting an energy threshold in the merging step to decide when the merging should be stopped. In this way, the number of clusters is related to the complexity of motions. As Fig.~\ref{fig:dynamicCluster} shows, when the motion is changed from one-arm-raising to two-arm-raising, the number of clusters is increased from 2 to 6 to represent the complex motion.

\end{document}